\documentclass[lettersize,journal]{IEEEtran}
\usepackage{amsmath,amsfonts}
\usepackage{bbm}
\usepackage{algorithmic}
\usepackage{algorithm}
\usepackage{array}
\usepackage[caption=false,font=normalsize,labelfont=sf,textfont=sf]{subfig}
\usepackage{textcomp}
\usepackage{stfloats}
\usepackage{url}
\usepackage{verbatim}
\usepackage{graphicx}
\usepackage{cite}
\usepackage{color}
\usepackage{multirow}
\usepackage{amssymb}
\usepackage[colorlinks=true, allcolors=blue]{hyperref}

\usepackage{colortbl}
\usepackage{xcolor}
\definecolor{background}{HTML}{e7f1f7}

\def\eg{{\it{e.g.}}}
\def\etal{{\it{et al.}}}
\def\ie{{\it{i.e.}}}
\begin{document}

\title{PIG: Prompt Images Guidance for Night-Time Scene Parsing}

\author{Zhifeng Xie, Rui Qiu, Sen Wang, Xin Tan*, Yuan Xie, Lizhuang Ma
\thanks{Manuscript received 28 Mar 2024; revised 19 May 2024.
This work was supported in part by the
National Natural Science Foundation of China under Grant 62302167, Grant U23A20343, Grant 62222602, and Grant 62176092; in part by  Chenguang Program of Shanghai Education Development Foundation and Shanghai Municipal Education Commission under Grant 23CGA34, Shanghai Sailing Program under Grant 23YF1410500; in part by the Natural Science Foundation of Chongqing, China, under Grant CSTB2023NSCQ-JQX0007 and Grant CSTB2023NSCQ-MSX0137; in part by the China Computer Federation (CCF)-Tencent Rhino-Bird Young Faculty Open Research Fund under Grant RAGR20230121; and in part by Shanghai Technical Service Center of Science and Engineering Computing, Shanghai University.
(*Corresponding author: Xin Tan.)}
\thanks{Zhifeng Xie is with the Department of Film and Television Engineering, Shanghai University, Shanghai 200072, China, and also with the Shanghai Key Laboratory of Computer Software Testing and Evaluating and the Shanghai Engineering Research Center of Motion Picture Special Effects, Shanghai 200072, China (e-mail: zhifeng\_xie@shu.edu.cn).
}
\thanks{Rui Qiu is with the Department of Film and Television Engineering, Shanghai University, Shanghai 200072, China (e-mail: ryan\_qr@shu.edu.cn).
}
\thanks{Sen Wang, Xin Tan and Yuan Xie are with the School of Computer Science and Technology, East China Normal University, Shanghai 200062, China, and also with Chongqing Institute of East China Normal University, Chongqing 401120, China (e-mail: wangsennn@foxmail.com; xtan@cs.ecnu.cn; yxie@cs.ecnu.edu.cn).}
\thanks{Lizhuang Ma is with the School of Computer Science and Technology, East China Normal University, Shanghai 200062, China, and also with the Department of Computer Science and Engineering, Shanghai Jiao Tong University, Shanghai 200240, China (e-mail: lzma@cs.ecnu.edu.cn).}
}
\markboth{Journal of \LaTeX\ Class Files,~Vol.~14, No.~8, August~2021}%
{Shell \MakeLowercase{\textit{et al.}}: A Sample Article Using IEEEtran.cls for IEEE Journals}


\maketitle

\begin{abstract}
Night-time scene parsing aims to extract pixel-level semantic information in night images, aiding downstream tasks in understanding scene object distribution. 
Due to limited labeled night image datasets, unsupervised domain adaptation (UDA) has become the predominant method for studying night scenes. 
UDA typically relies on paired day-night image pairs to guide adaptation, but this approach hampers dataset construction and restricts generalization across night scenes in different datasets. 
Moreover, UDA, focusing on network architecture and training strategies, faces difficulties in handling classes with few domain similarities.
In this paper, we leverage Prompt Images Guidance (PIG) to enhance UDA with supplementary night knowledge.
We propose a Night-Focused Network (NFNet) to learn night-specific features from both target domain images and prompt images.
To generate high-quality pseudo-labels, we propose Pseudo-label Fusion via Domain Similarity Guidance (FDSG).
Classes with fewer domain similarities are predicted by NFNet, which excels in parsing night features, while classes with more domain similarities are predicted by UDA, which has rich labeled semantics.
Additionally, we propose two data augmentation strategies: the Prompt Mixture Strategy (PMS) and the Alternate Mask Strategy (AMS), aimed at mitigating the overfitting of the NFNet to a few prompt images.
We conduct extensive experiments on four night-time datasets: NightCity, NightCity+, Dark Zurich, and ACDC. 
The results indicate that utilizing PIG can enhance the parsing accuracy of UDA.
The code is available at 
\href{https://github.com/qiurui4shu/PIG}{\textcolor{red}{https://github.com/qiurui4shu/PIG.}}

\end{abstract}


\section{Introduction}

\begin{figure}[!t]
    \centering
    \includegraphics[width=0.5\textwidth]{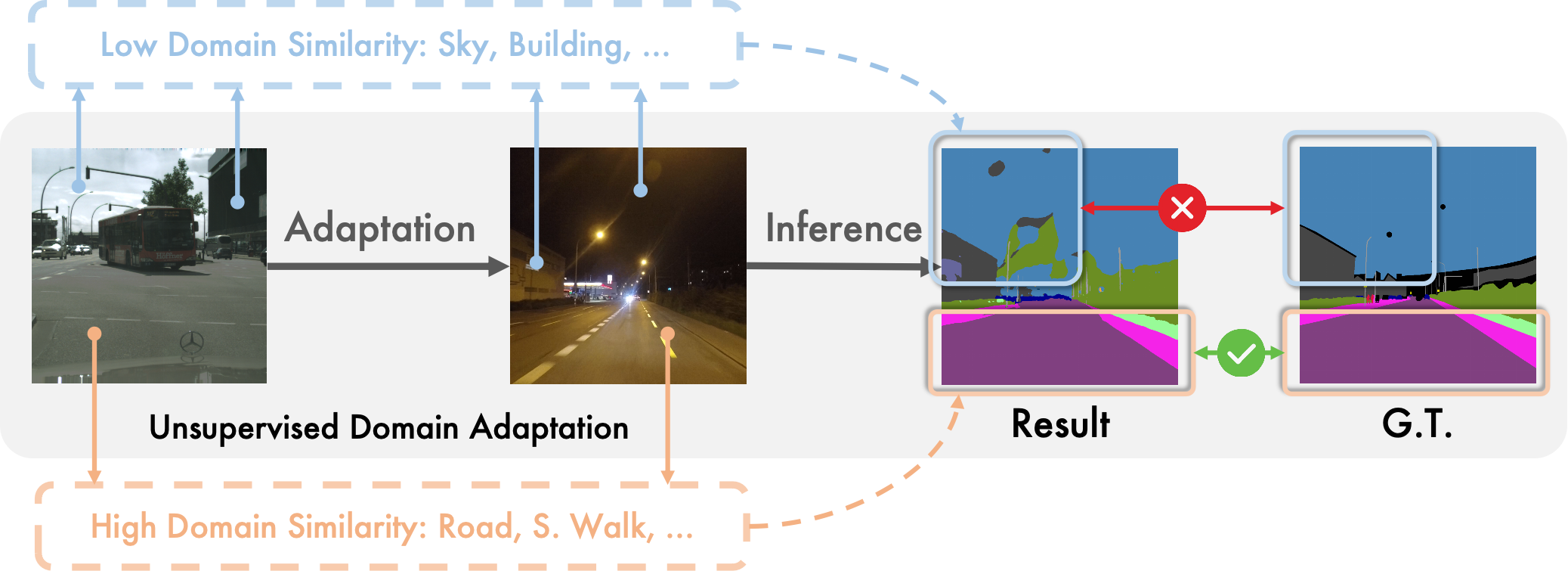}
    \vspace{-5mm}
    \caption{In unsupervised domain adaptation, the adaptation result tends to worsen as the similarity of object classes between the source and target domains decreases.}
    \vspace{-5mm}
    \label{fig:intro_similarity}
\end{figure}

\IEEEPARstart{S}{cene} parsing can extract pixel-level object attributes from real images to provide reliable analysis content for subsequent tasks, such as autonomous driving \cite{khoche2022semantic}, medical imaging \cite{lin2023rethinking}, and face recognition \cite{li2023robust}. 
Although there are many works focused on day-time scene parsing \cite{lu2023content},
night-time scene parsing (NTSP) \cite{tan2021night}  is still under-explored with greater challenges compared to day-time scenes due to varying degrees of exposure and low illumination interference, as well as dataset limitations.

Currently, unsupervised domain adaptation (UDA) methods \cite{gong2023continuous} are becoming popular to NTSP by transferring source-domain knowledge to the target domain since they do not require a large number of night-time labeled images.
For example, some methods \cite{hoyer2022daformer,hoyer2022hrda} draw on the traditional UDA structure to implement knowledge transferring directly,
but they did not explicitly model specific night-time features, resulting in unsatisfactory performance.
Meanwhile, a few UDA methods are specifically designed for NTSP, including DANNet \cite{wu2021dannet}, DANIA \cite{wu2021one}, and CCDistill \cite{gao2022cross}.
However, most of them require the paired day-night images for training the adaptation model, and strongly leverage the paired information as the priors. For example, these paired day-time and night-time images are  taken in the same location via the same viewpoint but only at different times; their static object distributions and scene structures are very similar. 
That is to say, these methods assume  very strong connections between day-time and night-time scenes are given, which is not practical.
Based on the discussion, a question arises: \textit{``can we establish a UDA model specifically designed for the NTSP task instead of learning from the paired day-to-night images?"}

Inspired by recent advancements in prompt engineering in large models, we introduce very few labeled night-time images (\ie, no more than 10 images) as prompts. 
These night-time images can be captured at any location, which means they are not paired with the given day-time images. 
The night-time images used for prompts have the following advantages: 1) they are easy to be obtained since there are several publicly available night-time semantic segmentation datasets, \eg, NightCity \cite{tan2021night}, Dark Zurich \cite{sakaridis2019guided}, and ACDC \cite{sakaridis2021acdc}, 2) they are generalized to the source day-time domain since no matching paired images are required, allowing this method to always function regardless of changes in the source images, and 3) the small number of night-times, at very low cost, can provide support for learning night-time features.

However, employing only a limited number of prompt images directly is difficult to affect the knowledge distribution supervised by a vast array of source domain images. 
Drawing insights from traditional UDA, it becomes apparent that domain similarity plays a pivotal role in the process of domain adaptation. 
As depicted in Fig. \ref{fig:intro_similarity}, the efficacy of domain adaptation hinges greatly upon the similarity between classes in the source and target domains. 
Favorable adaptation outcomes are observed when such similarity is high, whereas significant discrepancies may arise in prediction results when it is not. 
Therefore, we consider utilizing prompt images to guide object classes with low domain similarity, thereby alleviating the challenge of adapting these classes from the source domain to the target domain.

In this paper, we propose Prompt Images Guidance (PIG) to provide supplementary night knowledge for UDA. 
Prompt images and target domain images are trained together in the Night-Focused Network (NFNet).
The encoder and decoder in NFNet and UDA are kept consistent and trained simultaneously.
In order to take advantage of predictions of UDA and NFNet respectively, we first propose Pseudo-label \underline{F}usion via \underline{D}omain \underline{S}imilarity \underline{G}uidance (FDSG).
We use learned perceptual image patch similarity (LPIPS) \cite{zhang2018unreasonable} to evaluate the domain similarity of different classes in the source and target domains. 
In FDSG, fused pseudo-labels are generated by evaluating the LPIPS results of single-class images from the source domain and the target domain.
To avoid losing detailed information when fusing large objects, we design a small object preservation function to improve the accuracy of the fused labels.
Through FDSG, we can acquire high-quality pseudo-labels to supervise the training of both networks.
To enhance the learning of night features from a limited number of prompt images, we introduce two data augmentation strategies in NFNet: Prompt Mixture Strategy (PMS) and Alternate Mask Strategy (AMS).
By utilizing PMS to randomly concatenate target domain images and prompt images in a vertical segmentation manner, NFNet is empowered to acquire in-context inference for the same task and enhanced its robustness to image center-edge distribution.
The AMS works by randomly masking the image in blocks with a masking ratio.
The AMS largely eliminates spatial redundancy and prevents the network from losing detailed information due to masking.
We perform domain adaptation from the day-time datasets to the night-time datasets on extensive UDA methods and show different degrees of improvement.  
Our method can be embedded into extensive UDA methods as a flexible plug-in.
The main contributions are summarized in the following:

\begin{enumerate}
\item{We propose the Prompt Images Guidance (PIG) for NTSP. 
Training the Night-Focused Network (NFNet) exclusively on night images can improve the parsing accuracy for classes with night features.
To generate higher-quality pseudo-labels, we propose the Pseudo-label Fusion via Domain Similarity Guidance (FDSG), leveraging LPIPS to assess the domain similarity and guide the fusion of UDA and NFNet predicted pseudo-labels.
These fused pseudo-labels are then utilized for the self-supervision of both UDA and NFNet networks.
}
\item{To acquire comprehensive night knowledge from prompt images, we propose two data augmentation strategies during training NFNet: Prompt Mixture Strategy (PMS) and Alternate Mask Strategy (AMS). 
The PMS helps the NFNet learn night knowledge from a limited number of prompt images. 
The AMS eliminates spatial redundancy and prevents loss of detailed information, ensuring small objects are not neglected during network training.}
\item{Our method significantly enhances the parsing accuracy of UDA when adapting from day to night, and it can be seamlessly integrated into any UDA architecture.}
\end{enumerate}

\section{Related Work}
\subsection{Night-Time Scene Parsing}
Scene parsing involves obtaining pixel-level semantic segmentation results for night images, including 2D \cite{tan2021night} and 3D \cite{tan2023positive} scene parsing.
Compared to day-time scene parsing, night-time scene parsing faces greater challenges due to low light and over/under-exposure. 
Tan \etal \cite{tan2021night} are the first to develop a large-scale pixel-level labeled real night image dataset, NightCity/NightCity+. 
They propose an exposure-aware model to learn exposure features and improve the accuracy of scene parsing. 
Building upon this dataset, Xie \etal \cite{xie2023boosting} introduce a learnable frequency encoder that explores the frequency differences between day and night images. They fuse spatial domain and frequency domain information to extract features from night images.
NightLab \cite{deng2022nightlab} proposes a night segmentation framework that integrates light adaptation and segmentation modules at both the image and regional levels to enhance segmentation accuracy.

However, supervised models trained on the NighCity, the only labeled large-scale night dataset available, may not consistently deliver satisfactory results when applied to other night datasets. 
This is mainly because of domain gaps between different night datasets. 
We suggest using a few images from an existing labeled night dataset as prompt images when parsing other night scenes. 
This method enables the network to gain more accurate night knowledge and facilitates the flexible combination of images from various datasets.

\subsection{Unsupervised Domain Adaptation}
Unsupervised domain adaptation (UDA) is a subtask of transfer learning. 
UDA aims to learn from both heavily labeled source domain data and unlabeled target domain data in order to develop a model capable of solving tasks specific to the target domain. 
UDA encompasses various methods such as feature reconstruction \cite{ghifary2016deep}, adversarial training\cite{pei2018multi}, and distribution matching\cite{zhang2015deep}. 
UDA is particularly suitable for research areas where data scarcity poses a challenge. 
Due to the limited availability of labeled images in night scenes, researchers often resort to using labeled day-time datasets as source domains to address night-time scene parsing. 

On the one hand, some night datasets\cite{sakaridis2019guided,dai2018dark,sakaridis2021acdc} provide day-night image pairs, allowing the scene information in the day-time images to guide the parsing of night-time scenes.
Refign \cite{bruggemann2023refign} proposes an uncertainty-aware dense matching network to align day and night images and adaptively correct predicted pseudo-labels.
CCDistill \cite{gao2022cross} extracts the content and style knowledge contained in the features and calculate the degree of inherent or illumination difference between day and night images.
On the other hand, certain studies have shown improvements in parsing accuracy by optimizing model design and adjusting training settings.
DAFormer \cite{hoyer2022daformer} explores a network that is more suitable for UDA by improving the Transformer \cite{vaswani2017attention}. 
Additionally, it proposes three training strategies aimed at significantly enhancing the parsing results for rare and thing classes.
HRDA \cite{hoyer2022hrda} combines the advantages of high and low resolution crops to capture long-range context dependencies while preserving fine segmentation details.

After mutually constraining day-night image pairs, the pseudo-labels generated during self-supervision can be effectively corrected. 
However, this approach inevitably increases the cost of studying night-time scene parsing. 
Specifically, when exploring new night scenes, creating a dataset of day-night image pairs is much more demanding compared to a dataset consisting solely of night images. 
The methods for optimizing UDA architecture are typically universal, leading to the frequent oversight of potential night features in labeled night images.
Prompt image guidance can address the limitations of the aforementioned research methods. 
By incorporating a few night prompt images, the labeling costs can be reduced. 
Additionally, we aim to maximize night feature extraction from the prompt images and utilize them to guide UDA training, thereby enhancing domain adaptation in night-time scene parsing.

\subsection{Prompt Learning}
Prompt learning is a method initially employed in natural language processing (NLP) to align the objective of a downstream task with the objective of a pre-trained model. 
This approach addresses the challenge posed by the excessive number of parameters in the pre-trained model \cite{brown2020language}, making it impractical for fine-tuning. 
Numerous studies \cite{schick2021exploiting} have demonstrated that prompt learning can significantly enhance NLP model accuracy.
In computer vision (CV), various image recognition tasks often require fine-tuning of the entire large model.
The introduction of prompt learning offers a promising direction to address this issue effectively.
Prompt learning has been successfully applied to a wide range of CV tasks, including classification \cite{jia2022visual}, anomaly detection \cite{li2024promptad}, continual learning \cite{wang2022learning}, multi-modal learning \cite{zhou2022learning}, domain adaptation \cite{ge2022domain}, and more. 
Models like CLIP \cite{radford2021learning} use contrastive learning to align the feature space of text and images, enabling visual models to transfer based on different text prompts \cite{gal2022stylegan}.
Large-scale segmentation models, such as SAM \cite{kirillov2023segment} and Painter \cite{wang2023images}, integrate various tasks including semantic segmentation, object detection, panoramic segmentation, and depth estimation into a single comprehensive model using prompt coding or prompt images.

Prompt learning is commonly utilized by large models to fine-tune downstream tasks, yet it is often disregarded when it comes to smaller models. 
This is mainly because smaller models usually concentrate on specific tasks that don't necessitate prompts for task completion. 
Moreover, the limited and fixed data used in prompt learning fails to meet the requirement for data diversity in deep learning. 
However, in UDA, we believe there is still value in exploring the potential of prompt learning. 
In the transition from day images to night images, the fewer the domain similarities within the same class, the more difficult it becomes for the class to gather knowledge from day images that can be effectively applied for night parsing. 
Prompt images can offer additional guidance for classes with minimal domain similarities, thereby facilitating smoother domain adaptation for the model.

\begin{figure*}[!t]
    \centering
    \includegraphics[width=1\textwidth]{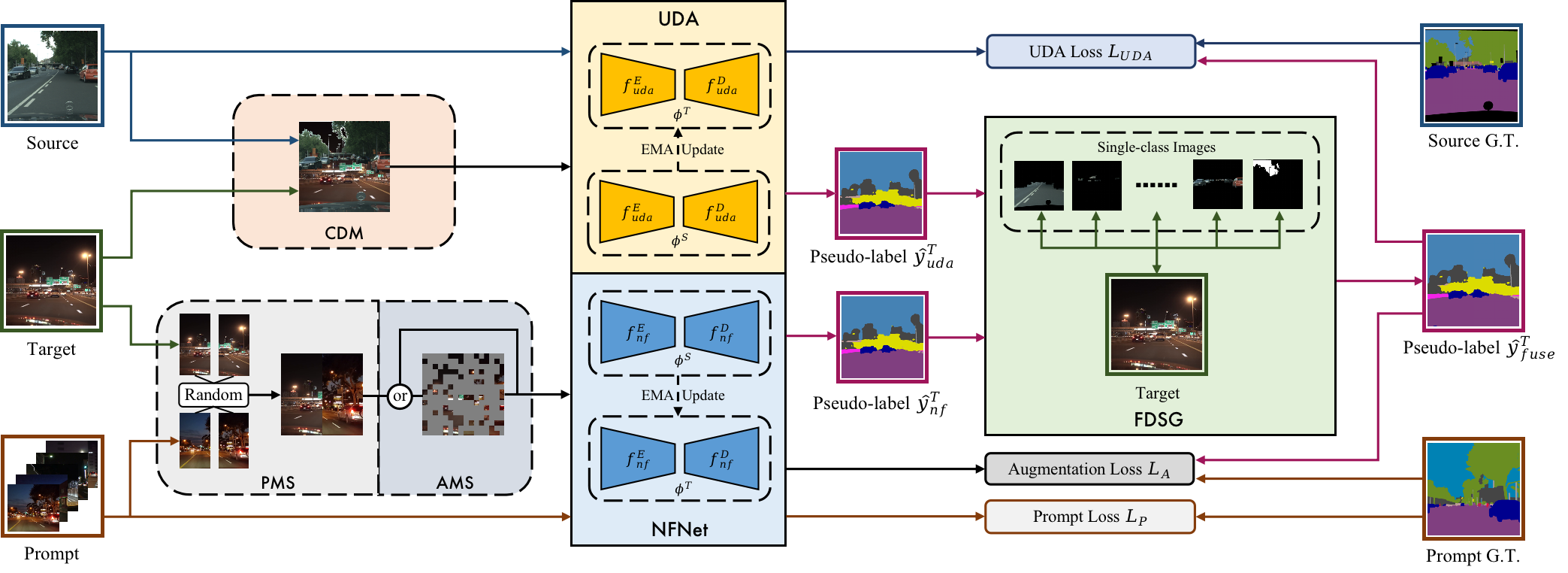}
    \vspace{-7mm}
    \caption{The training pipeline of the Prompt Images Guidance (PIG). 
    The datasets include labeled day images from the source domain, a labeled set of prompt images, and unlabeled night images from the target domain. 
    UDA trains on both day and night images. 
    When processing target domain images, a Cross-Domain Mixed (CDM) data augmentation method is commonly employed for both night and day images.
    NFNet exclusively trains on night images, where the input night images are combined with the prompt images using the Prompt Mixture Strategy (PMS) and the Alternate Mask Strategy (AMS).
    The predictions $\hat{y}^{T}_{uda}$ and $\hat{y}^{T}_{nf}$, from both UDA and NFNet are then fed into the FDSG module.
    FDSG is detailed in Fig. \ref{fig:FDGG}.
    Depending on the ordering of the FDSG results, $\hat{y}^{T}_{uda}$ and $\hat{y}^{T}_{nf}$ are fused with different class weights to produce the pseudo-label $\hat{y}^{T}_{fuse}$.
    Finally, $\hat{y}^{T}_{fuse}$ is utilized in UDA to calculate UDA losses $L_{UDA}$ with the ground truth from the source domain, while in NFNet, $\hat{y}^{T}_{fuse}$ participates in the augmentation losses $L_{A}$ and prompt losses $L_{P}$ along with the ground truth from the prompt images.
    }
    \vspace{-5mm}
    \label{fig:pipeline}
\end{figure*}

\section{The Proposed Method}

\subsection{Components of the Prompt Images Guidance}
\subsubsection{UDA and NFNet Architecture}
In the UDA in Fig. \ref{fig:pipeline}, the primary objective is to adapt features of the source domain $D_{S} = \{(x_{S},y_{S})\}_{N_{S}}$ to the target domain $D_{T} = \{x_{T}\}_{N_{T}}$.
This is achieved through a combination of supervised training on the source domain and self-supervised training on the target domain.
Noting that only the images $x$ in the source domain have ground truth $y$.
In the present research \cite{hoyer2022daformer,hoyer2022hrda,hoyer2023mic}, UDA training networks are commonly built upon the Transformer \cite{vaswani2017attention} architecture, comprising an encoder $f_{uda}^{E}$ for generating feature maps and a decoder $f_{uda}^{D}$ designed to handle the scene parsing task.
Furthermore, to ensure the generation of stable pseudo-labels during the training process, the UDA architecture incorporates a teacher-student network. 
The teacher network, which remains detached from gradient track, updates its network parameters $\phi^{T}$ using exponential moving average (EMA) based on the parameters $\phi^{S}$ of the student network:
\begin{equation}
    \label{ema}
    \phi^{T}_{i+1} \leftarrow  \alpha \phi^{T}_{i} + (1-\alpha) \phi^{S}_{i}
    \text{,}
\end{equation}
where $\alpha$ denotes the EMA factor and $i$ denotes the number of training iterations.

Nevertheless, training a network on images from two different domains simultaneously can introduce a challenge in learning the features specific to the target domain. 
For instance, in the day-time domain, the sky and vegetation exhibit prominent color and edge features, whereas in the night-time domain, these features become less discernible due to low light conditions. 
Owing to the substantial difference between the two domains for certain classes, the network's understanding of these classes becomes unclear, making it difficult to achieve satisfactory parsing results for both day and night images.

To address the aforementioned challenge, we propose the Night-Focused Network that leverages prompt images $D_{P} = \{(x_{P},y_{P})\}_{N_{P}}$. 
NFNet maintains an identical architecture $f_{nf}^{E},f_{nf}^{D}$ to the encoder-decoder used in UDA and retains the teacher-student network training mode.
During training, NFNet and UDA are synchronized. 
Prompt images refer to a small set of labeled night images, which can be sourced from the target domain or added as supplementary night scenes. 
NFNet exclusively receives prompt images and night images from the target domain, avoiding interference from day-time images. 
This enables NFNet to better learn night-specific features. 
However, NFNet's comprehensive parsing ability for night scenes is not stronger than that of UDA, mainly due to the lack of a diverse range of labeled image supervision available in the source domain.

\subsubsection{Pseudo-label Fusion via Domain Similarity Guidance}
The pseudo-labels generated by the teacher network play a crucial role in enabling the network to adapt from the source domain to the target domain in UDA. 
High-quality pseudo-labels can effectively mitigate confirmation bias \cite{arazo2020pseudo} during network self-training. 
However, when training the model with both day-time and night-time images, the pseudo-labels generated by the model may be influenced by the day-time knowledge, deviating from the actual scene distribution, especially for classes exhibiting few domain similarities. 
Considering that the NFNet, trained exclusively on night images, demonstrates superior parsing accuracy for classes with night features, we design the fusion of UDA and NFNet generated pseudo-labels. 
This fusion approach aims to generate higher-quality pseudo-labels, which can then supervise the training of the model.

\begin{figure}[!t]
    \centering
    \includegraphics[width=0.5\textwidth]{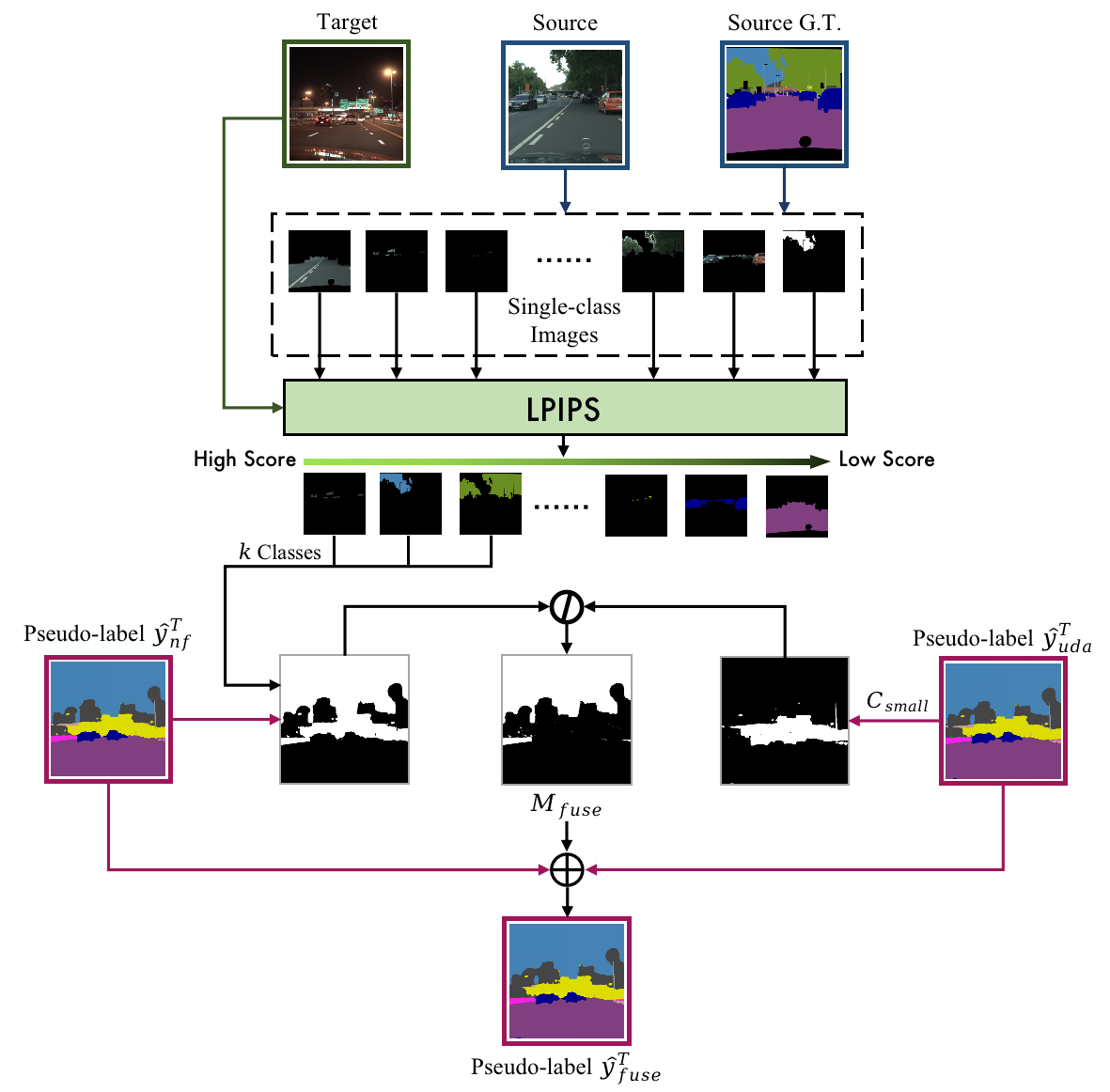}
    \vspace{-7mm}
    \caption{Pseudo-label Fusion via Domain Similarity Guidance (FDSG). 
    The input day image generates a single-class image that contains only one class, based on the corresponding ground truth.
    These single-class images are sequentially fed into the LPIPS module alongside the target domain night image.
    The network evaluates single-class images and night image, producing an evaluation result for each single-class image. 
    A higher evaluation result indicates a greater difficulty in adapting the class during this iteration. 
    We select the first $k$ classes to guide the prediction $\hat{y}^{T}_{nf}$ of NFNet, generating a mask that contains $k$ classes. 
    Additionally, we employ UDA to predict small objects in $\hat{y}^{T}_{uda}$, generating a small object mask. 
    The two masks are then subtracted from each other to obtain the fused mask $M_{fuse}$. 
    Finally, fusion pseudo-label $\hat{y}^{T}_{fuse}$ is generated by combining $\hat{y}^{T}_{nf}$, $\hat{y}^{T}_{uda}$, and $M_{fuse}$ according to formula \eqref{eq_fuse}.
    }
    \vspace{-5mm}
    \label{fig:FDGG}
\end{figure}

We propose a pseudo-label fusion via domain similarity guidance, as illustrated in Fig. \ref{fig:FDGG}.
When the target domain image $x_{T}$ is fed into both teacher encoder-decoder $f^{D} \circ f^{E}$, UDA and NFNet generate two corresponding predictions. This operation can be expressed as:
\begin{equation}
    \hat{y}^{T}_{uda} = \sigma (f^{D}_{uda} \circ f^{E}_{uda} (x_{T}))
\end{equation}
\begin{equation}
    \hat{y}^{T}_{nf} = \sigma (f^{D}_{nf} \circ f^{E}_{nf} (x_{T}))
    \text{,}
\end{equation}
where $\sigma (\cdot)$ denotes the softmax operator. 
Each iteration of the network is considered as a domain adaptation process.
However, if the domain similarity between specific classes in the two images is few, it becomes difficult for this domain adaptation to learn the common features of these classes.
For these classes, it is more appropriate to use NFNet for prediction, as it excels at capturing night-specific features.
Therefore, we need to evaluate the domain similarity between classes in day and night images.

We note that Learned Perceptual Image Patch Similarity (LPIPS) \cite{zhang2018unreasonable} is commonly employed as a metric to evaluate the similarity between images generated by Generative Adversarial Networks (GANs) and reference images.
Hence, we employ the LPIPS metric as an indicator to evaluate the domain similarity between day-time domain and night-time domain.
Based on the labeled image $y_{S}$, we isolate each class within the day image $x_{S}$ and reconstruct $c$ single-class images $x_{c}$:
\begin{equation}
    x_{c} = [ \; y_{S} = c \;] \odot x_{S}
    \text{,}
\end{equation}
where $c$ denotes the class contained in the $y_{S}$ and the $\odot$ is element-wise multiplication.
Classes that do not appear in the day-time image will be excluded in this iteration.
Subsequently, we compute the LPIPS scores between these single-class images and the night-time images:
\begin{equation}
    g_{c} = LPIPS(x_{c},x_{T})
    \text{,}
\end{equation}
where $LPIPS(\cdot,\cdot)$ denotes the result of inputting two images into the pre-trained LPIPS model.
The higher the $g_{c}$, the fewer the domain similarity.
By arranging $g_{c}$ in descending order
\begin{equation}
    R_{c} = Rank_{\downarrow} (\; \{ g_{c} \}_{n} \;)
    \text{,}
\end{equation}
we select the first $k$ classes for NFNet parsing, while the remaining classes are assigned to UDA.
Due to the abundance of supervision information for small objects in UDA, and the high similarity in shape and texture features of these objects across different domains, UDA is better for parsing small objects.
Additionally, the fusion of pseudo-labels can potentially lead to a loss of details.
Therefore, the predictions of small object classes from UDA are retained when $\hat{y}^{T}_{uda}$ and $\hat{y}^{T}_{nf}$ are fused:
\begin{equation}
    M_{fuse} = [\; \hat{y}^{T}_{nf} = R_{[\; c \in k  \;]} \;] - [\; \hat{y}^{T}_{uda} = C_{small} \;]
\end{equation}
\begin{equation}
    \label{eq_fuse}
    \hat{y}^{T}_{fuse} = M_{fuse} \odot \hat{y}^{T}_{nf} + \thicksim M_{fuse} \odot \hat{y}^{T}_{uda}
    \text{,}
\end{equation}
where $-$ denotes the minus and $\thicksim$ denotes the binary inverse operator.
Fused pseudo-label $\hat{y}^{T}_{fuse}$ are utilized for the self-supervised training of UDA and NFNet.

\subsection{Data Augmentation Strategies}
\subsubsection{Prompt Mixture Strategy}
Cross-domain mixed (CDM) source domain image and target domain image training network is a common data augmentation method in UDA. 
During the image mixing process, a random selection is made from the ground truth classes in the source domain image, and half of the object classes are retained. 
Subsequently, the retained image is overlaid onto the target domain image, resulting in a mixed image that combines both day-time and night-time scenes, as illustrated in Fig. \ref{fig:pipeline}.
This mixed image can be considered as a transitional domain, reducing the complexity of adapting from the source domain to the target domain. 
Moreover, the cross-domain mixture approach also provides abundant supervision information for the network, particularly for parsing small objects in the target domain.

When training NFNet, our aim is to effectively utilize the supervision information provided by the prompt images. 
However, due to the significantly smaller number of prompt images compared to the source domain images, the cross-domain mixture may not be the optimal mixing method for NFNet. 
Firstly, the limited number of prompt images results in a high repetition rate in the content of the generated mixture of images. 
Consequently, the network is exposed to redundant information. 
Additionally, the presence of similar images in the mixture can lead to overfitting on the prompt images. 
Secondly, when mixing images based on the distribution of ground truth classes, there is a high likelihood of having fewer object classes in the mixture image compared to the real image. 
As a result, the diversity of the training data is reduced, impeding the network to learn from a wide range of object classes. 
Although this limitation can be mitigated by leveraging a large amount of data in the source domain during cross-domain mixture, it becomes more pronounced when working with small amounts of data.

To enable a reasonable mixture of prompt images $x_{P}$ and target domain images $x_{T}$, we propose a prompt mixture strategy, as shown in Fig. \ref{fig:pipeline}.
In this strategy, we do not consider the object type as the criterion for mixing the two image types. 
Inspired by Painter \cite{wang2023images}, we view the prompt image as a scene task prompt and the target domain image as a related task to be accomplished.
To mix these two image types, we begin by generating a prompt mask, denoted as $M_{p} \in \mathbb{R}^{H \times W}$, which has the same size as the input image. 
Given the clear hierarchical distribution of object classes in night scenes along the horizontal axis (e.g., the arrangement of sky, building, and road from top to bottom), we employ a left-right division scheme when mixing $x_{P}$ and $x_{T}$. 
This ensures that the number of object classes does not decrease significantly. 
We randomly assign the value 1 to one side of the mask $M_{p}$ and 0 to the other:
\begin{equation}
    \label{p_mask}
    a \in \{0, 1\}, \quad
    M_{p}^{H \times W} =
    \begin{cases}
        a & 0<w \leq \frac{W}{2} \\
        1-a & \frac{W}{2} < w \leq W
    \end{cases}
    \text{.}
\end{equation}
Subsequently, we generate the task prompt image, denoted as
\begin{equation}
    \label{x_pm}
    x_{pm} = M_{p} \odot x_{P} + (1 - M_{p}) \odot x_{T}
    \text{,}
\end{equation}
where the $\odot$ is element-wise multiplication.

After training on prompt mixture images, NFNet is able to achieve in-context inference within the same domain and improve its robustness to the distribution of the center and edge of the image. 
Unlike cross-domain mixture, the prompt mixture doesn't necessitate a substantial amount of data to enable the model to comprehend parsing tasks within a given scene and yield superior performance.

\subsubsection{Alternate Mask Strategy}
NFNet, trained on a limited set of labeled prompt images, effectively learned that accurate supervision information is often minimal.
To maximize the utilization of this supervised information, we propose employing Masked Autoencoder (MAE) \cite{he2022masked} to enable the NFNet to learn the inference between neighboring objects, as shown in Fig. \ref{fig:pipeline}.
Due to its advantages of simplicity, high efficiency, and high performance, MAE has been extensively utilized in Transformer-based research. 
Similarly, we generate a patch mask $M_{a} \in \mathbb{R}^{H \times W}$ with a masking ratio of $r$. 
We divided $M_{a}$ into patches with size $p \times q$. Each patch can be represented by $N_{i,j}$. 
We randomly assign a set of values $z_{i,j} \sim \mathcal{U}(0,1)$ conforming uniform distribution to patches. 
Cover patches in $M_{a}$:
\begin{equation}
    \label{ams}
    M_{a[ip:(i+1)p,jq:(j+1)q]} = [\; z_{i,j} > r \;]
    \text{.}
\end{equation}
When the task prompt image requires masking, we perform element-wise multiplication between the $M_{a}$ and the task prompt image and subsequently input it into NFNet for training.

While MAE can effectively reduce spatial redundancy and enhance the model's capacity to infer adjacent objects from context, a high mask rate can potentially lead to the complete loss of rare small objects in the image. 
Considering that the early participation of rare classes during model training significantly contributes to the model's parsing ability \cite{hoyer2022daformer}, we propose the alternate mask strategy:
\begin{equation}
    \label{x_am}
    x_{am} = (MOD(t,\frac{1}{\beta}) + MOD(t,\frac{1}{\beta}) M_{a}) \odot x_{pm}
    \text{,}
\end{equation}
where $t$ denotes the number of training iterations and $\beta$ denotes the participation rate of $M_{a}$ in training. $MOD$ refers to the modulus operation in mathematics.

By employing the alternate mask strategy, the model ensures that rare class small objects are not disregarded during the process of learning to reconstruct predictions.

\begin{figure}[!t]
    \centering
    \includegraphics[width=0.38\textwidth]{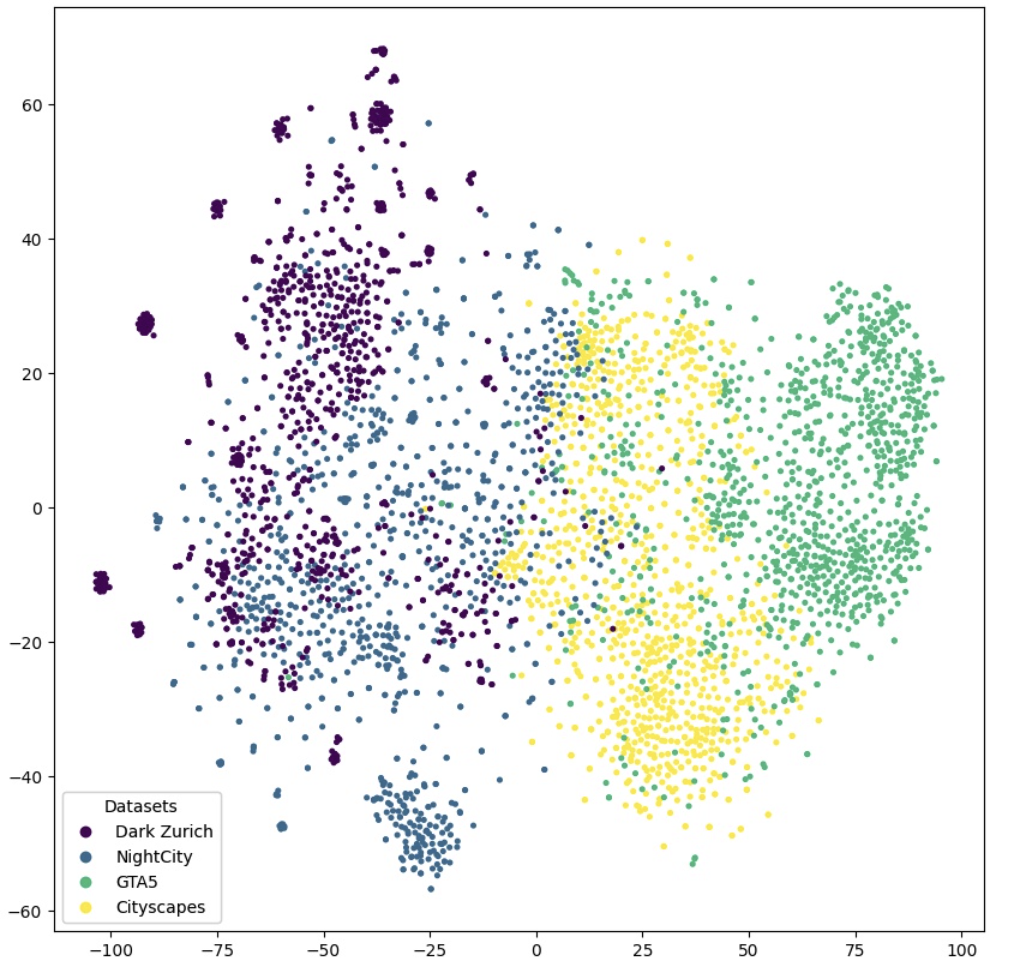}
    \vspace{-2mm}
    \caption{T-SNE visualization of datasets for virtual city, day-time, and night-time domains.}
    \vspace{-2mm}
    \label{fig:t-sne}
\end{figure}

\begin{table}[!t]
    \centering
    \caption{Comparison of SOTA methods in unsupervised domain adaptation and full supervision. Full-supervise (FS) means that only the labeled target domain images participate in the training. The results are from \cite{hoyer2023mic,wang2023internimage,wei2023disentangle}.}
    \renewcommand{\arraystretch}{1.3}
    \begin{tabular}{c|ccc}
    \hline
        Datasets & UDA & FS & UDA/FS\\
        \hline
        GTA5 $\rightarrow$ Cityscapes & 75.9 mIoU & 86.1 mIoU & 88.3\% \\
        Cityscapes $\rightarrow$ NightCity & 41.2 mIoU & 61.2 mIoU & 67.3\% \\
    \hline
    \end{tabular}
    \vspace{-3mm}
    \label{tab:Domain_adaptation}
\end{table}

\subsection{Loss Function}
We utilize the categorical cross-entropy loss to regulate the training of the model. 
Due to variations in UDA architectures, the loss function in UDA is uniformly denoted as $\mathcal{L}_{UDA}$. 
It is crucial to highlight that in UDA, when employing pseudo-labels for self-supervision of target domain images, we substitute the pseudo-labels with our $\hat{y}^{T}_{fuse}$.
NFNet includes prompt image supervision loss
\begin{equation}
    \mathcal{L}_{P} =  \sum^{H \times W}_{i,j=1} \sum^{C}_{c=1} y^{i,j,c}_{P} \log \sigma (f^{D}_{nf} \circ f^{E}_{nf} (x_{P}))^{i,j,c}
    \text{,}
\end{equation}
and self-supervision loss of images after data augmentation
\begin{equation}
    \mathcal{L}_{A} =  \sum^{H \times W}_{i,j=1} \sum^{C}_{c=1} \hat{y}^{i,j,c}_{A} \log \sigma (f^{D}_{nf} \circ f^{E}_{nf} (x_{am}))^{i,j,c}
    \text{,}
\end{equation}
where $\hat{y}_{A}$ is the pseudo-label for $x_{am}$ generated by equations \eqref{x_pm} \eqref{x_am}, but replace $x_{P},x_{T}$ with $y_{P},\hat{y}^{T}_{fuse}$ in \eqref{x_pm}. The overall loss $\mathcal{L}$ is the weighted sum of the presented loss component
\begin{equation}
    \mathcal{L} = \lambda_{1} \mathcal{L}_{UDA} + \lambda_{2} \mathcal{L}_{P} + \lambda_{3} \mathcal{L}_{A}
    \text{,}
    \label{eqn_15}
\end{equation}
where $\lambda$ denotes a hyperparameter that regulates the weight of the loss.

\subsection{Day-to-Night Domain Adaptation}

Currently, unsupervised domain adaptation tasks in scene parsing mainly concentrate on adapting from virtual city domains to real day-time domains. 
In order to assess the disparities in data between day to night domain adaptation and general tasks, we employ T-SNE to reduce dimensionality and visualize the datasets for the virtual city, day-time, and night-time domains, as depicted in Fig. \ref{fig:t-sne}.
We randomly sample 1000 images from each of the following datasets: GTA5\cite{richter2016playing} (virtual city domain), Cityscapes\cite{cordts2016cityscapes} (day-time domain), NightCity\cite{tan2021night} and Dark Zurich\cite{sakaridis2019guided} (night-time domain).
Furthermore, we investigate the state-of-the-art methods for domain adaptation tasks along with target domain full-supervised tasks, which are summarized in Tab. \ref{tab:Domain_adaptation}.

In Fig. \ref{fig:t-sne}, we can observe that the data distribution of the virtual city domain and the day-time domain is highly concentrated, and the distance between the data points of these two domains is relatively close.
This indicates that the data characteristics of these two domains are quite similar, and the domain similarity between them is higher.
On the contrary, the data distribution in the night-time domain is more scattered, indicating that the features of night images are not as concentrated as those of day images. 
Even within the same night domain, the features of the images can vary considerably.
Additionally, the distance between the data points in the night-time domain and those in the day-time domain is greater than the distance between the data points in the virtual city domain and the day-time domain.
This suggests that adapting from day-time to night-time is more challenging.
The findings in Tab. \ref{tab:Domain_adaptation} indicate that utilizing only the supervision information from the virtual city domain allows achieving 88.3\% accuracy on the full supervision of the day-time domain.
However, when utilizing the day-time domain for supervision, the accuracy achieved on the night-time domain is only 67.3\%.
This observation further reinforces the idea that adapting from day to night domains is more challenging than the general unsupervised domain adaptation task.

Due to the substantial domain difference between day and night, parsing models face challenges when classifying objects with different features in different domains under the same class.
However, training the network exclusively on target domain images and combining classes of varying similarities with different weights from UDA and NFNet predictions makes it possible to alleviate the blurring caused by feature inconsistency.

\section{Experiments}
\subsection{Datasets and Implementation Details}
\textit{Datasets:} We utilize the day-time dataset Cityscapes \cite{cordts2016cityscapes} as the source domain and adapt to the target night scenes in different night-time datasets, including NightCity \cite{tan2021night}, NightCity+ \cite{deng2022nightlab}, Dark Zurich \cite{sakaridis2019guided} and ACDC\cite{sakaridis2021acdc}. 
The prompt images are selected from NightCity and Dark Zurich.

\begin{enumerate}
    \item {\textit{Cityscapes:} It consists of 2048$\times$1024 resolution day-time images captured from real-world of European street scenes, comprising 2975 training images with ground truth, as well as 500 validation images.}
    \item {\textit{NightCity:} It comprises 4,297 annotated 1024$\times$512 images, with 2,998 images allocated for training and 1,299 for validation. The labeling standards employed in this dataset are identical to Cityscapes.}
    \item {\textit{NightCity+:} Building upon NightCity, this dataset enhances the image resolution of the validation set to 2048$\times$1024, refines the labeling with greater detail, and rectifies certain errors found in the original data.}
    \item {\textit{Dark Zurich:} It encompasses a total of 2,416 night-time images, 2,920 twilight images, and 3,041 day-time images. These images are unlabeled and possess a resolution of 1920$\times$1080. Additionally, the dataset includes 201 annotated night-time images, with 50 allocated for validation and 151 for testing.}
    \item {\textit{ACDC:} It consists of 4,006 city images under adverse conditions, including 1,000 foggy images, 1,006 night-time images, 1,000 rainy images, and 1,000 snowy images. All images have a resolution of 1920$\times$1080 and have been fine annotated. For our purposes, we exclusively utilize the night-time images from the dataset and do not employ any corresponding ground truth.} 
\end{enumerate}

\textit{Network Architecture:}
DAFormer \cite{hoyer2022daformer} is a typical method in UDA that focuses on designing training strategies, while HRDA \cite{hoyer2022hrda} is a typical method to improve UDA by optimizing UDA architecture.
Therefore, we investigate the impact of PIG in the context of these two types of methods.
The entire network is designed within the MMSegmentation\footnote{https://github.com/open-mmlab/mmsegmentation} framework. 
The DAFormer utilizes a Transformer-based architecture, comprising a MiT-B5 \cite{xie2021segformer} encoder and a context-aware feature fusion decoder. 
Building upon the DAFormer, the HRDA method uses an extra lightweight SegFormer MLP decoder \cite{xie2021segformer}. 
The teacher network and NFNet share the same architecture as the student network. 
In the LPIPS model, we use Alexnet \cite{krizhevsky2017imagenet} for feature extraction.

\textit{Training:} In model training, we set a worker, a batch size of 2 (4 workers, a batch size of 2 in DAFormer \cite{hoyer2022daformer}), 40k training iterations, SGD with a learning rate of $2.5\times10^{-3}$.
In the UDA, following HRDA \cite{hoyer2022hrda}, we set AdamW \cite{loshchilovdecoupled} with a learning rate of $6\times10^{-5}$ for the encoder and $6\times10^{-4}$ for the decoder, a warmup learning rate, a loss weight of 1, an EMA factor of 0.999, data augmentation of DACS \cite{tranheden2021dacs}.
Additionally, we set a masking ratio of 0.7, a participation rate of 0.5, a classes rank $k$ of 4,  hyperparameters $\lambda_{1}=\lambda_{2}=\lambda_{3}=1$.
In the experiments where the number of prompt images is not discussed, we specify that the default number of prompt images is 10.

\textit{Inference:} In the inference stage, test images are directly processed through the UDA’s student network to obtain predictions, without involving NFNet in the inference process.

\begin{table*}[!t]
    \centering
    \caption{Comparison results of PIG and UDA methods on 4 benchmarks.
    $^{\star}$ indicates that the method utilizes day-night image pairs as an additional resource.
    }
    \renewcommand{\arraystretch}{1.3}
    \resizebox{1\linewidth}{!}{
    \begin{tabular}{lccccccccccccccccccc|c}
    \hline
         Method & Road & S.walk & Build. & Wall & Fence & Pole & Tr.Light & Sign & Veget. & Terrain & Sky & Person & Rider & Car & Truck & Bus & Train & M.bike & Bike & mIoU \\
         \hline
         \multicolumn{21}{c}{ \textbf{Cityscapes $\rightarrow$  NightCity (Test)}}\\
         \hline
         ADVENT \cite{vu2019advent} & 83.10 & 27.20 & 61.10 & 4.50 & 6.30 & 16.70 & 11.70 & 24.50 & 16.60 & 10.00 & 1.40 & 37.40 & \underline{9.80} & 62.90 & 14.30 & 24.50 & 9.00 & 7.30 & 22.00 & 24.43
         \\
         \hline
         DAFormer \cite{hoyer2022daformer} & 85.80 & 25.70 & 70.40 & \textbf{19.10} & 10.30 & \underline{31.20} & 18.40 & 28.40 & 16.90 & 4.00 & 15.50 & \underline{49.60} & 9.30 & 69.80 & 61.20 & 50.70 & 20.10 & 16.50 & 26.20 & 33.10
         \\
         \cellcolor{background}PIG (DAFormer) & \cellcolor{background}86.86 & \cellcolor{background}\underline{27.22} & \cellcolor{background}\textbf{75.88} & \cellcolor{background}\underline{16.05} & \cellcolor{background}11.10 & \cellcolor{background}\textbf{31.99} & \cellcolor{background}20.00 & \cellcolor{background}41.34 & \cellcolor{background}\textbf{46.97} & \cellcolor{background}\underline{11.38} & \cellcolor{background}\textbf{82.06} & \cellcolor{background}48.07 & \cellcolor{background}6.83 & \cellcolor{background}\underline{73.57} & \cellcolor{background}62.84 & \cellcolor{background}\underline{61.15} & \cellcolor{background}\textbf{36.04} & \cellcolor{background}20.51 & \cellcolor{background}18.67 & \cellcolor{background}\underline{40.98} (+7.88)
         \\
         HRDA \cite{hoyer2022hrda} & \textbf{88.60} & \textbf{35.50} & 67.80 & 16.40 & \textbf{19.00} & 29.20 & \textbf{23.70} & \underline{41.80} & 16.80 & \textbf{17.10} & 13.10 & \textbf{52.20} & \textbf{14.30} & \textbf{77.20} & \textbf{65.90} & \textbf{61.90} & 29.20 & \underline{28.30} & \underline{31.40} & 38.38
         \\
         \cellcolor{background}PIG (HRDA) & \cellcolor{background}\underline{87.10} & \cellcolor{background}26.90 & \cellcolor{background}\underline{75.10} & \cellcolor{background}14.30 & \cellcolor{background}\underline{12.20} & \cellcolor{background}26.80 & \cellcolor{background}\underline{23.00} & \cellcolor{background}\textbf{44.70} & \cellcolor{background}\underline{46.80} & \cellcolor{background}11.10 & \cellcolor{background}\underline{81.80} & \cellcolor{background}46.50 & \cellcolor{background}2.80 & \cellcolor{background}69.80 & \cellcolor{background}\underline{65.50} & \cellcolor{background}52.20 & \cellcolor{background}\underline{33.50} & \cellcolor{background}\textbf{30.90} & \cellcolor{background}\textbf{32.20} & \cellcolor{background}\textbf{41.20} (+2.28)
         \\
         \hline
         \multicolumn{21}{c}{ \textbf{Cityscapes $\rightarrow$  NightCity+ (Test)}}\\
         \hline
         ADVENT \cite{vu2019advent} & 84.10 & 28.10 & 62.00 & 4.40 & 6.30 & 16.70 & 11.70 & 27.00 & 17.30 &
         13.10 & 1.30 & 40.10 & 16.80 & 64.70 & 14.00 & 25.00 & 8.40 & 6.40 & 23.10 & 25.67
         \\
         \hline
         DAFormer \cite{hoyer2022daformer} & 87.30 & 26.20 & 70.60 & \underline{19.30} & 10.50 & \textbf{30.50} & 22.90 & 33.70 & 17.90 & 11.10 & 15.90 & 53.80 & \underline{17.10} & 72.50 & \underline{61.20} & 54.80 & 20.70 & 19.10 & \underline{32.30} & 35.65
         \\
         \cellcolor{background}PIG (DAFormer) & \cellcolor{background}\underline{88.00} & \cellcolor{background}27.50 & \cellcolor{background}\underline{74.90} & \cellcolor{background}\textbf{20.30} & \cellcolor{background}\textbf{18.50} & \cellcolor{background}24.40 & \cellcolor{background}14.30 & \cellcolor{background}33.20 & \cellcolor{background}\textbf{48.60} & \cellcolor{background}\underline{24.80} & \cellcolor{background}\underline{79.70} & \cellcolor{background}42.60 & \cellcolor{background}1.50 & \cellcolor{background}71.10 & \cellcolor{background}53.60 & \cellcolor{background}\underline{60.90} & \cellcolor{background}\textbf{34.80} & \cellcolor{background}8.20 & \cellcolor{background}13.90 & \cellcolor{background}38.93 (+3.28)
         \\
         HRDA \cite{hoyer2022hrda} & \textbf{90.00} & \textbf{36.60} & 67.20 & 16.20 & \underline{16.70} & 25.00 & \underline{27.10} & \textbf{49.30} & 17.40 & \textbf{25.10} & 13.50 & \textbf{60.20} & \textbf{27.20} & \textbf{80.80} & \textbf{64.80} & \textbf{66.30} & \underline{34.40} & \textbf{28.30} & \textbf{42.00} & \underline{41.46}
         \\
         \cellcolor{background}PIG (HRDA) & \cellcolor{background}87.10 & \cellcolor{background}\underline{30.20} & \cellcolor{background}\textbf{76.40} & \cellcolor{background}19.20 & \cellcolor{background}14.20 & \cellcolor{background}\underline{29.80} & \cellcolor{background}\textbf{27.80} & \cellcolor{background}\underline{48.70} & \cellcolor{background}\underline{42.60} & \cellcolor{background}16.30 & \cellcolor{background}\textbf{81.80} & \cellcolor{background}\underline{55.40} & \cellcolor{background}9.80 & \cellcolor{background}\underline{79.80} & \cellcolor{background}60.40 & \cellcolor{background}51.10 & \cellcolor{background}26.60 & \cellcolor{background}\underline{20.80} & \cellcolor{background}16.70 & \cellcolor{background}\textbf{41.82} (+0.36)
         \\
         \hline
         \multicolumn{21}{c}{ \textbf{Cityscapes $\rightarrow$  Dark Zurich (Test)}}\\
         \hline
         DACS \cite{tranheden2021dacs} & 83.10 & 49.10 & 67.40 & 33.20 & 16.60 & 42.90 & 20.70 & 35.60 & 31.70 & 5.10 & 6.50 & 41.70 & 18.20 & 68.80 & 76.40 & 0.00 & 61.60 & 27.70 & 10.70 & 36.70
         \\
         MGCDA$^{\star}$ \cite{sakaridis2020map} & 80.30 & 49.30 & 66.20 & 7.80 & 11.00 & 41.40 & 38.90 & 39.00 & 64.10 & 18.00 & 55.80 & 52.10 & 53.50 & 74.70 & 66.00 & 0.00 & 37.50 & 29.10 & 22.70 & 42.50
         \\
         CDAda$^{\star}$ \cite{xu2021cdada} & 90.50 & 60.60 & 67.90 & 37.00 & 19.30 & 42.90 & 36.40 & 35.30 & 66.90 & 24.40 & 79.80 & 45.40 & 42.90 & 70.80 & 51.70 & 0.00 & 29.70 & 27.70 & 26.20 & 45.00
         \\
         DANNet$^{\star}$ \cite{wu2021dannet} & 90.40 & 60.10 & 71.00 & 33.60 & 22.90 & 30.60 & 34.30 & 33.70 & 70.50 & 31.80 & 80.20 & 45.70 & 41.60 & 67.40 & 16.80 & 0.00 & 73.00 & 31.60 & 22.90 & 45.20
         \\
         DANIA$^{\star}$ \cite{wu2021one} & 91.50 & 62.70 & \underline{73.90} & 39.90 & \textbf{25.70} & 36.50 & 35.70 & 36.20 & 71.40 & 35.30 & \textbf{82.20} & 48.00 & 44.90 & 73.70 & 11.30 & 0.10 & 64.30 & 36.70 & 22.70 & 47.00
         \\
         CCDistill$^{\star}$ \cite{gao2022cross} & 89.60 & 58.10 & 70.60 & 36.60 & 22.50 & 33.00 & 27.00 & 30.50 & 68.30 & 33.00 & 80.90 & 42.30 & 40.10 & 69.40 & 58.10 & 0.10 & 72.60 & 47.70 & 21.30 & 47.50
         \\
         IR$^{2}$F-RMM \cite{gong2023continuous} & \textbf{94.70} & \textbf{75.10} & 73.20 & \textbf{44.40} & \textbf{25.70} & \textbf{60.60} & 39.00 & \underline{47.40} & 70.20 & \textbf{41.60} & 77.30 & \textbf{62.40} & 55.50 & \textbf{86.40} & 55.50 & \textbf{20.00} & \textbf{92.00} & \textbf{55.30} & \textbf{42.80} & \underline{58.90}
         \\
         \hline
         DAFormer \cite{hoyer2022daformer} & \underline{93.50} & 65.50 & 73.30 & 39.40 & 19.20 & 53.30 & 44.10 & 44.00 & 59.50 & 34.50 & 66.60 & 53.40 & 52.70 & 82.10 & 52.70 & 9.50 & 89.30 & 50.50 & 38.50 & 53.80
         \\
         \cellcolor{background}PIG (DAFormer) & \cellcolor{background}92.40 & \cellcolor{background}65.90 & \cellcolor{background}\textbf{75.10} & \cellcolor{background}42.10 & \cellcolor{background}18.80 & \cellcolor{background}53.80 & \cellcolor{background}30.10 & \cellcolor{background}41.10 & \cellcolor{background}\textbf{73.90} & \cellcolor{background}31.40 & \cellcolor{background}\underline{82.10} & \cellcolor{background}55.60 & \cellcolor{background}49.30 & \cellcolor{background}80.00 & \cellcolor{background}\underline{59.30} & \cellcolor{background}15.40 & \cellcolor{background}90.00 & \cellcolor{background}47.50 & \cellcolor{background}36.10 & \cellcolor{background}54.72 (+0.92)
         \\
         HRDA \cite{hoyer2022hrda} & 90.40 & 56.30 & 72.00 & 39.50 & 19.50 & \underline{57.80} & \textbf{52.70} & 43.10 & 59.30 & 29.10 & 70.50 & \underline{60.00} & \textbf{58.60} & \underline{84.00} & 75.50 & 11.20 & \underline{90.50} & \underline{51.60} & \underline{40.90} & 55.90
         \\
         \cellcolor{background}PIG (HRDA) & \cellcolor{background}91.80 & \cellcolor{background}\underline{73.30} & \cellcolor{background}73.40 & \cellcolor{background}\underline{43.60} & \cellcolor{background}20.80 & \cellcolor{background}57.70 & \cellcolor{background}\underline{49.40} & \cellcolor{background}\textbf{54.30} & \cellcolor{background}\underline{71.70} & \cellcolor{background}\underline{38.10} & \cellcolor{background}80.50 & \cellcolor{background}58.70 & \cellcolor{background}\underline{56.50} & \cellcolor{background}82.40 & \cellcolor{background}\textbf{80.70} & \cellcolor{background}\underline{17.30} & \cellcolor{background}89.90 & \cellcolor{background}41.80 & \cellcolor{background}40.40 & \cellcolor{background}\textbf{59.06} (+3.16)
         \\
         \hline
         \multicolumn{21}{c}{ \textbf{Cityscapes $\rightarrow$  ACDC-Night (Test)}}
         \\
         \hline
         CDAda$^{\star}$ \cite{xu2021cdada} & 74.70 & 29.50 & 49.40 & 17.10 & 12.60 & 31.00 & 38.20 & 30.00 & 48.00 & 22.80 & 0.20 & 47.00 & 25.40 & 63.80 & 12.80 & 46.10 & 23.10 & 24.70 & 24.60 & 32.70
         \\
         MGCDA$^{\star}$ \cite{sakaridis2020map} & 74.50 & 52.50 & 69.40 & 7.70 & 10.80 & 38.40 & 40.20 & 43.30 & 61.50 & 36.30 & 37.60 & 55.30 & 25.60 & 71.20 & 10.90 & 46.40 & 32.60 & 27.30 & 33.80 & 40.80
         \\
         DANNet$^{\star}$ \cite{wu2021dannet} & 90.70 & 61.20 & 75.60 & 35.90 & \underline{28.80} & 26.60 & 31.40 & 30.60 & \textbf{70.80} & 39.40 & \underline{78.70} & 49.90 & 28.80 & 65.90 & 24.70 & 44.10 & 61.10 & 25.90 & 34.50 & 47.60
         \\
         DANIA$^{\star}$ \cite{wu2021one} & 91.00 & 60.90 & \underline{77.70} & 40.30 & \textbf{30.70} & 34.30 & 37.90 & 34.50 & \underline{70.00} & 37.20 & \textbf{79.60} & 45.70 & 32.60 & 66.40 & 11.10 & 37.00 & 60.70 & 32.60 & 37.90 & 48.30
         \\
         IR$^{2}$F-RMM \cite{gong2023continuous} & \textbf{92.80} & \underline{64.80} & 74.50 & \underline{42.40} & 15.00 & \underline{51.70} & 36.70 & \underline{52.40} & 66.60 & \textbf{46.70} & 62.70 & \underline{64.10} & 36.30 & 80.30 & \textbf{59.80} & \textbf{72.10} & \textbf{87.70} & 32.00 & \underline{50.50} & \textbf{57.30}
         \\
         \hline
         DAFormer \cite{hoyer2022daformer} & 74.80 & 58.60 & 72.70 & 30.60 & 19.80 & 38.70 & 15.70 & 37.20 & 49.10 & 43.30 & 45.20 & 56.90 & 25.40 & 68.70 & 14.30 & 40.20 & 82.70 & 30.60 & 44.10 & 44.70
         \\
         \cellcolor{background}PIG (DAFormer) & \cellcolor{background}\underline{92.40} & \cellcolor{background}63.50 & \cellcolor{background}75.20 & \cellcolor{background}36.40 & \cellcolor{background}19.40 & \cellcolor{background}46.70 & \cellcolor{background}38.30 & \cellcolor{background}41.00 & \cellcolor{background}47.40 & \cellcolor{background}38.60 & \cellcolor{background}61.50 & \cellcolor{background}56.20 & \cellcolor{background}27.70 & \cellcolor{background}70.00 & \cellcolor{background}43.50 & \cellcolor{background}45.20 & \cellcolor{background}81.50 & \cellcolor{background}\textbf{38.00} & \cellcolor{background}45.70 & \cellcolor{background}50.96 (+6.26)
         \\
         HRDA \cite{hoyer2022hrda} & 87.30 & 46.20 & 76.00 & 35.70 & 17.50 & \textbf{52.00} & \textbf{50.30} & \textbf{53.60} & 53.10 & 44.00 & 41.70 & \textbf{64.80} & \underline{40.90} & \underline{76.30} & \underline{49.10} & \underline{64.80} & \underline{83.10} & \underline{36.00} & \textbf{51.50} & 53.90
         \\
         \cellcolor{background}PIG (HRDA) & \cellcolor{background}91.90 & \cellcolor{background}\textbf{70.80} & \cellcolor{background}\textbf{81.30} & \cellcolor{background}\textbf{44.70} & \cellcolor{background}13.90 & \cellcolor{background}50.50 & \cellcolor{background}\underline{44.60} & \cellcolor{background}51.80 & \cellcolor{background}68.80 & \cellcolor{background}\underline{45.70} & \cellcolor{background}78.60 & \cellcolor{background}62.10 & \cellcolor{background}\textbf{42.20} & \cellcolor{background}\textbf{76.60} & \cellcolor{background}41.70 & \cellcolor{background}63.60 & \cellcolor{background}78.70 & \cellcolor{background}25.20 & \cellcolor{background}48.40 & \cellcolor{background}\underline{56.91} (+3.01)
         \\
         \hline
    \end{tabular}
    }
    \label{tab:result}
    \vspace{-3mm}
\end{table*}

\subsection{Experimental Results}
\subsubsection{Comparison with UDA Methods}
To validate the effectiveness of our approach, we conduct a comparison between PIG and unsupervised domain adaptation methods on four night datasets: NightCity, NightCity+, Dark Zurich, and ACDC. 
The source domain image utilized for training comes from the labeled training set in Cityscapes, while the target domain image is obtained from the training set in the night dataset.
The evaluation results are derived from the test set of the night dataset.
It's worth noting that in DAFormer, the trained images is downsampled to a quarter of original size.

\textit{Methods for Comparisons:}
We comprehensively compare unsupervised domain adaptation methods, particularly those not relying on day-night image pairs.
The evaluated UDA methods include DAFormer\cite{hoyer2022daformer}, HRDA\cite{hoyer2022hrda}, ADVENT\cite{vu2019advent}, DACS\cite{tranheden2021dacs}, and IR$^{2}$F-RMM \cite{gong2023continuous}. 
Notably, our proposed approach, PIG, leverages DAFormer and HRDA as part of its UDA component. 
Furthermore, we compare UDA methods that utilize day-night image pairs, namely MGCDA\cite{sakaridis2020map}, CDAda\cite{xu2021cdada}, DANNet\cite{wu2021dannet}, DANIA\cite{wu2021one}, and CCDistill\cite{gao2022cross}.

\textit{Quantitative Comparison:}
The experimental results are presented in Tab. \ref{tab:result}, showcasing the performance of different methods. 
Compared to DAFormer, our proposed approach, PIG (DAFormer), exhibits improvements of +7.88 mIoU, +3.28 mIoU, +0.92 mIoU, and +6.26 mIoU on the night datasets NightCity, NightCity+, Dark Zurich, and ACDC, respectively. 
Similarly, when compared to HRDA, PIG (HRDA) demonstrates improvements of +2.28 mIoU, +0.36 mIoU, +3.16 mIoU, and +3.01 mIoU, respectively. 
Notably, PIG achieves significant improvements in accuracy for both vegetation and sky classes. 
Furthermore, PIG outperforms other methods utilizing day-night image pairs, suggesting that UDA can achieve higher-quality pseudo-labels through its own architectural optimization.
This indicates that correcting night predictions doesn't necessarily rely on matching day-time scene images.
Although all of our baseline results align with those published in the corresponding paper, they do not accurately represent our local experimental conditions. 
As a result, Tab. \ref{tab:result} shows a small increase of 0.36 mIoU.

\begin{figure*}
    \centering
        \begin{minipage}{0.16\linewidth}
            \centerline{\footnotesize{Image}}
        \end{minipage}
        \begin{minipage}{0.16\linewidth}
            \centerline{\footnotesize{G.T.}}
        \end{minipage}
        \begin{minipage}{0.16\linewidth}
            \centerline{\footnotesize{DAFormer\cite{hoyer2022daformer}}}
        \end{minipage}
        \begin{minipage}{0.16\linewidth}
            \centerline{\footnotesize{PIG (DAFormer)}}
        \end{minipage}
        \begin{minipage}{0.16\linewidth}
            \centerline{\footnotesize{HRDA\cite{hoyer2022hrda}}}
        \end{minipage}
        \begin{minipage}{0.16\linewidth}
            \centerline{\footnotesize{PIG (HRDA)}}
        \end{minipage}
        \vskip 5pt
        \begin{minipage}{0.16\linewidth}
            \includegraphics[width=2.95cm]{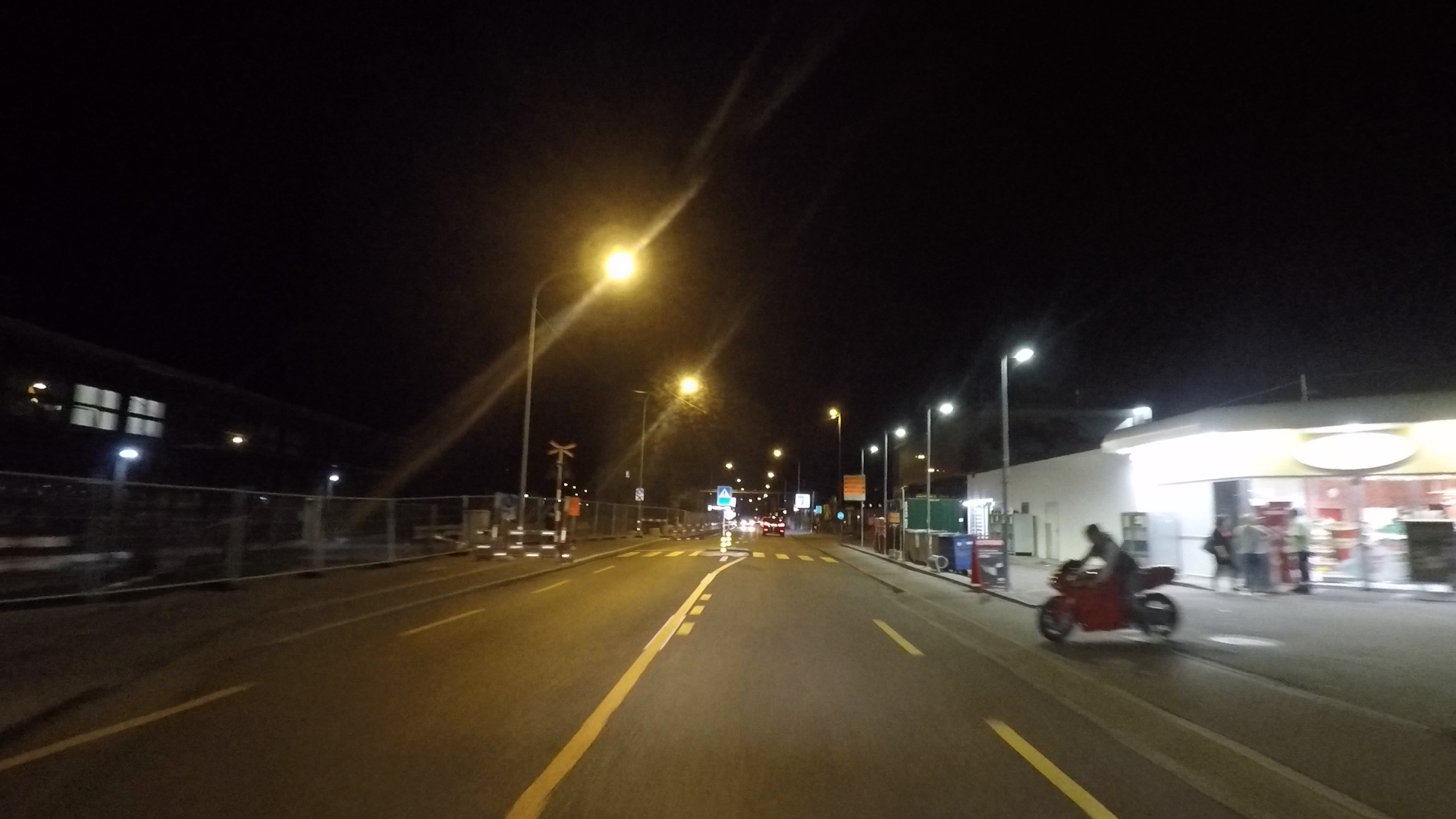}
        \end{minipage}
        \begin{minipage}{0.16\linewidth}
            \includegraphics[width=2.95cm]{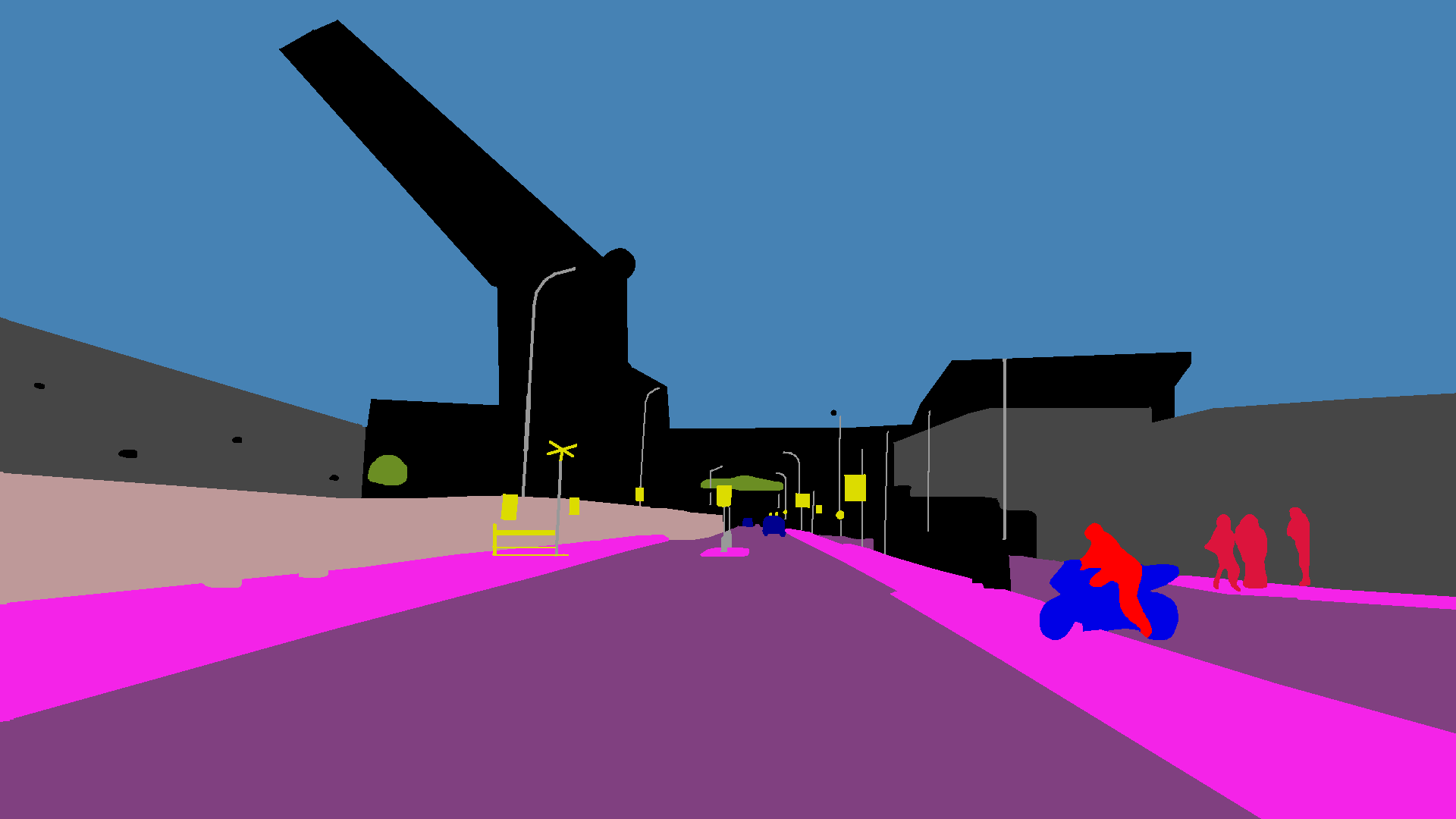}
        \end{minipage}
        \begin{minipage}{0.16\linewidth}
            \includegraphics[width=2.95cm]{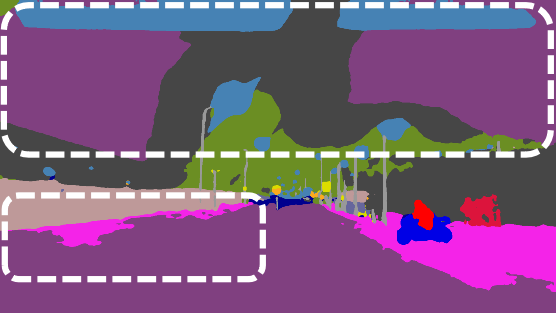}
        \end{minipage}
        \begin{minipage}{0.16\linewidth}
            \includegraphics[width=2.95cm]{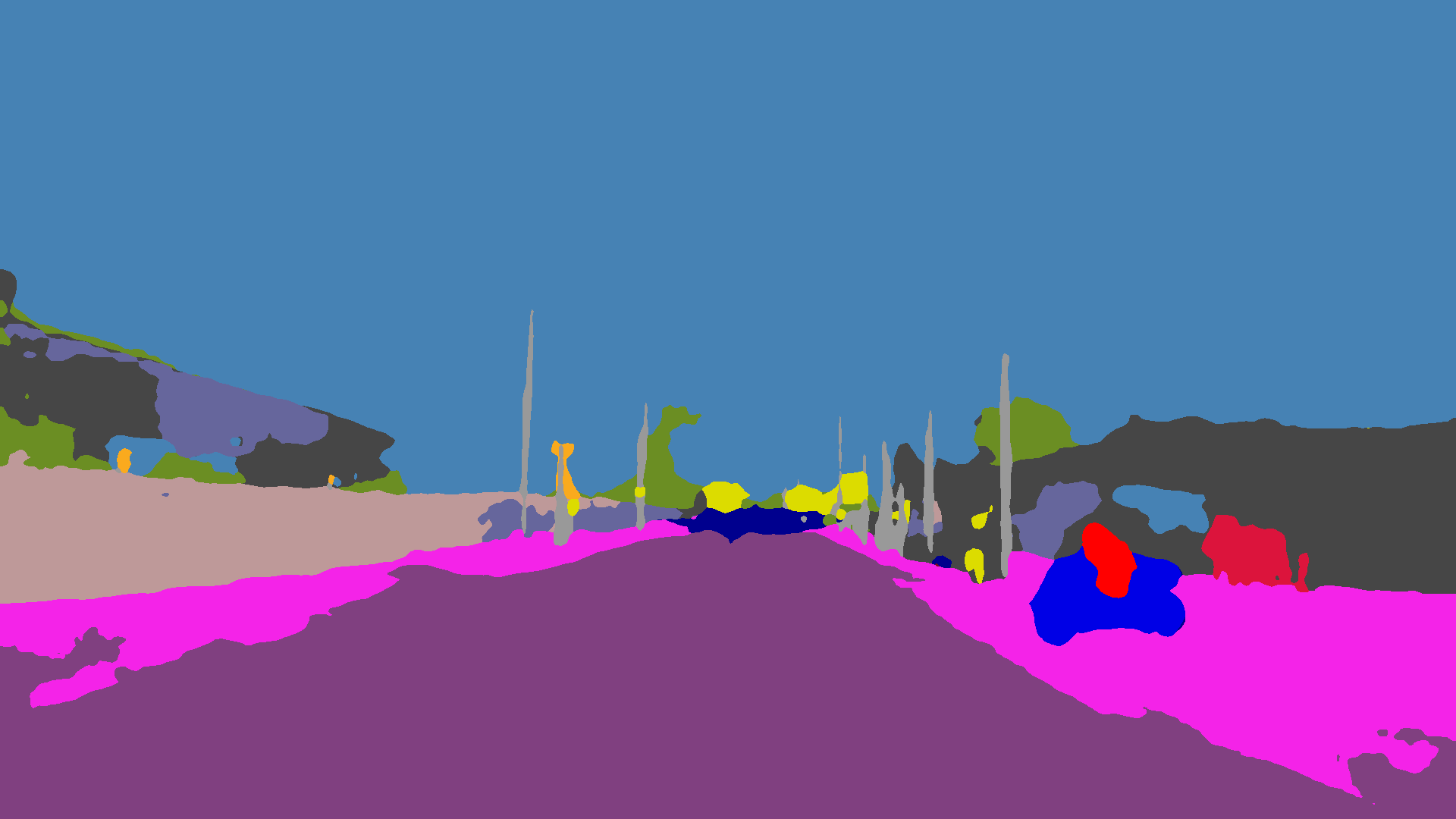}
        \end{minipage}
        \begin{minipage}{0.16\linewidth}
            \includegraphics[width=2.95cm]{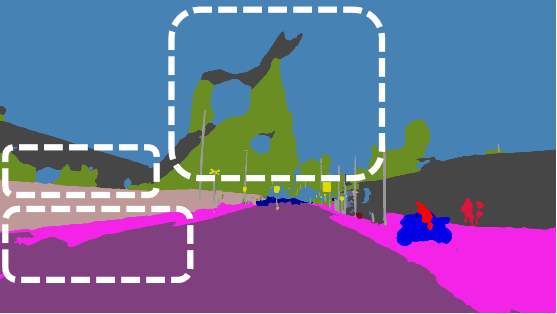}
        \end{minipage}
        \begin{minipage}{0.16\linewidth}
            \includegraphics[width=2.95cm]{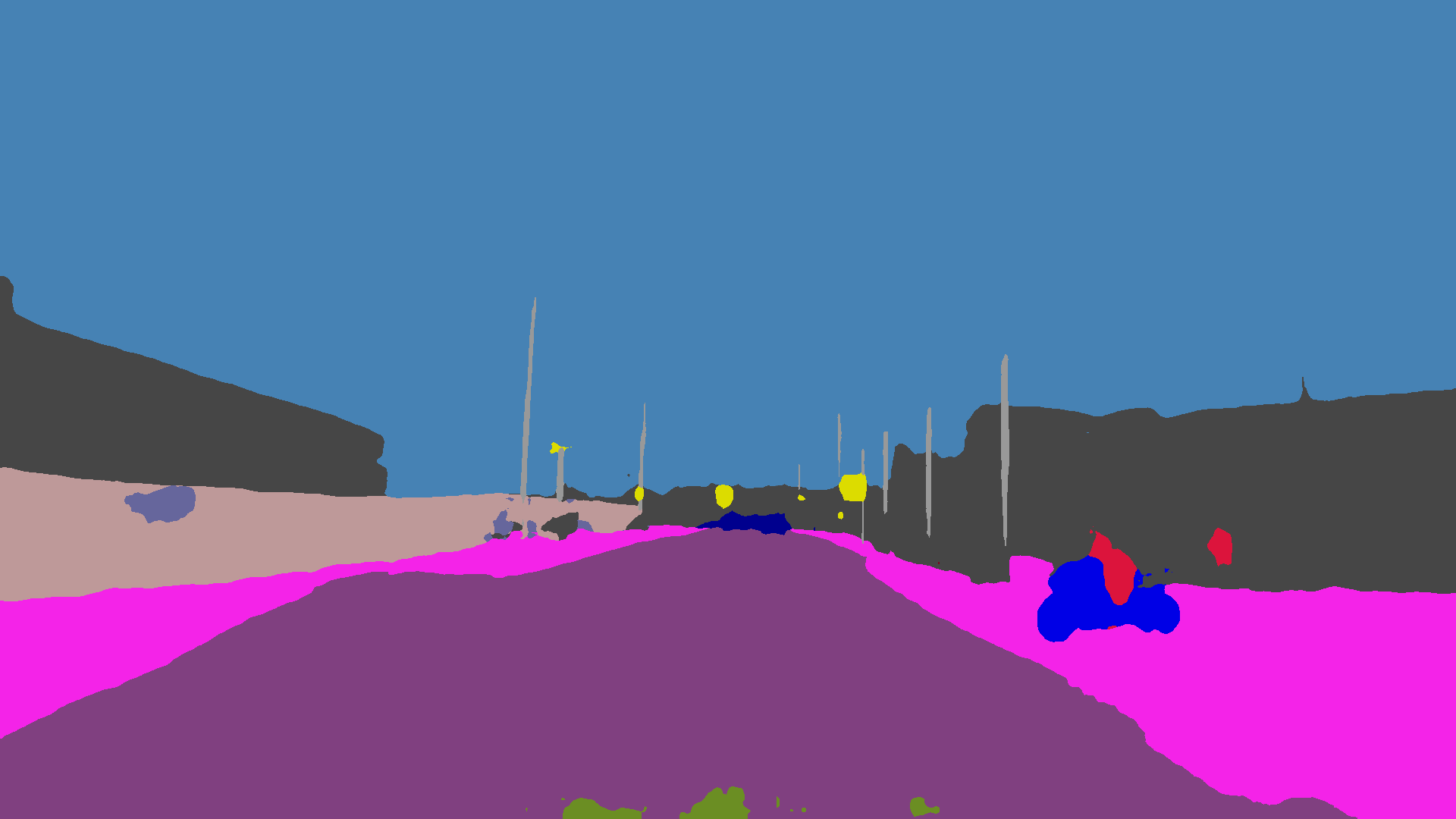}
        \end{minipage}
        \vskip 2pt
        \begin{minipage}{0.16\linewidth}
            \includegraphics[width=2.95cm]{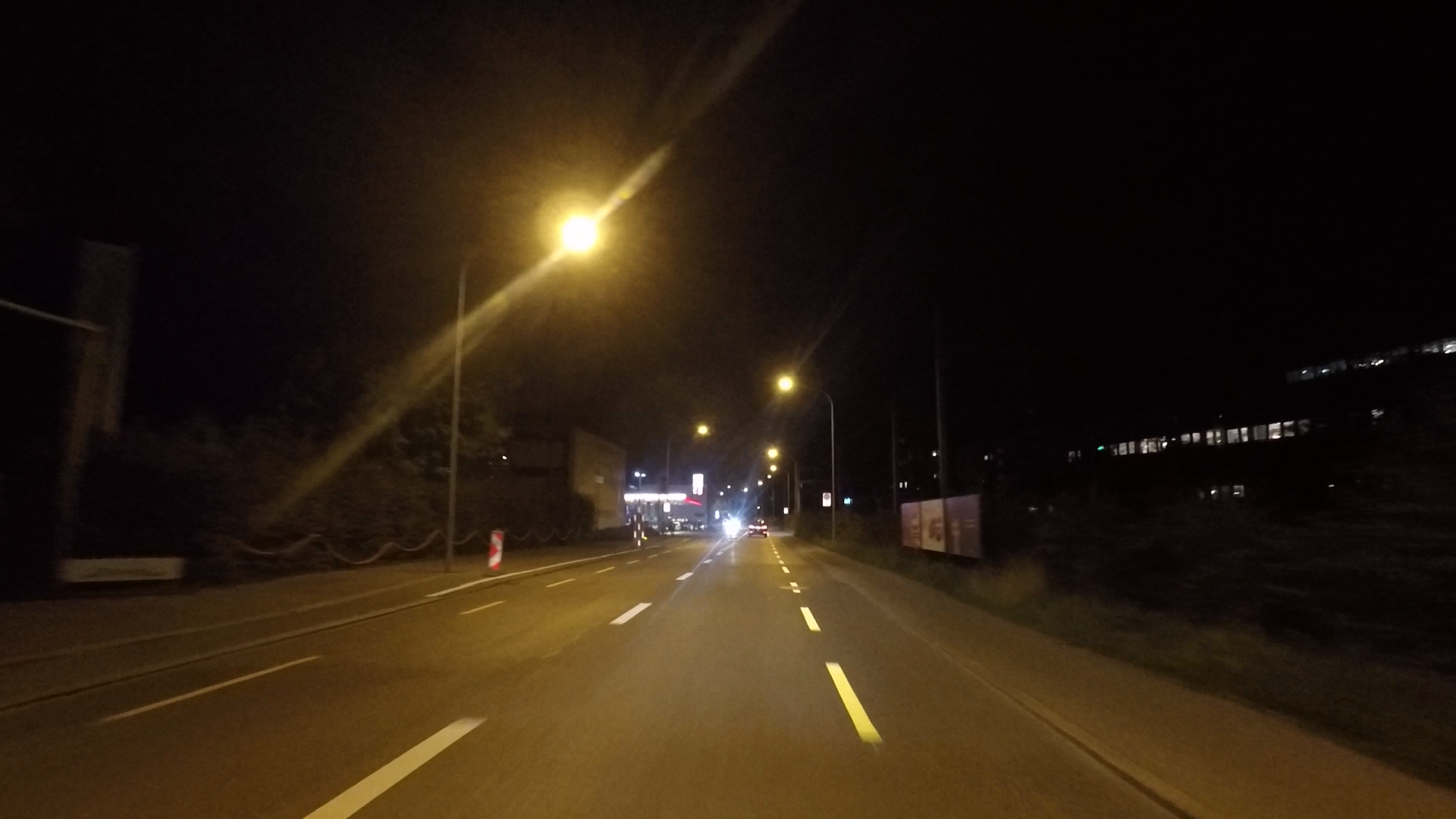}
        \end{minipage}
        \begin{minipage}{0.16\linewidth}
            \includegraphics[width=2.95cm]{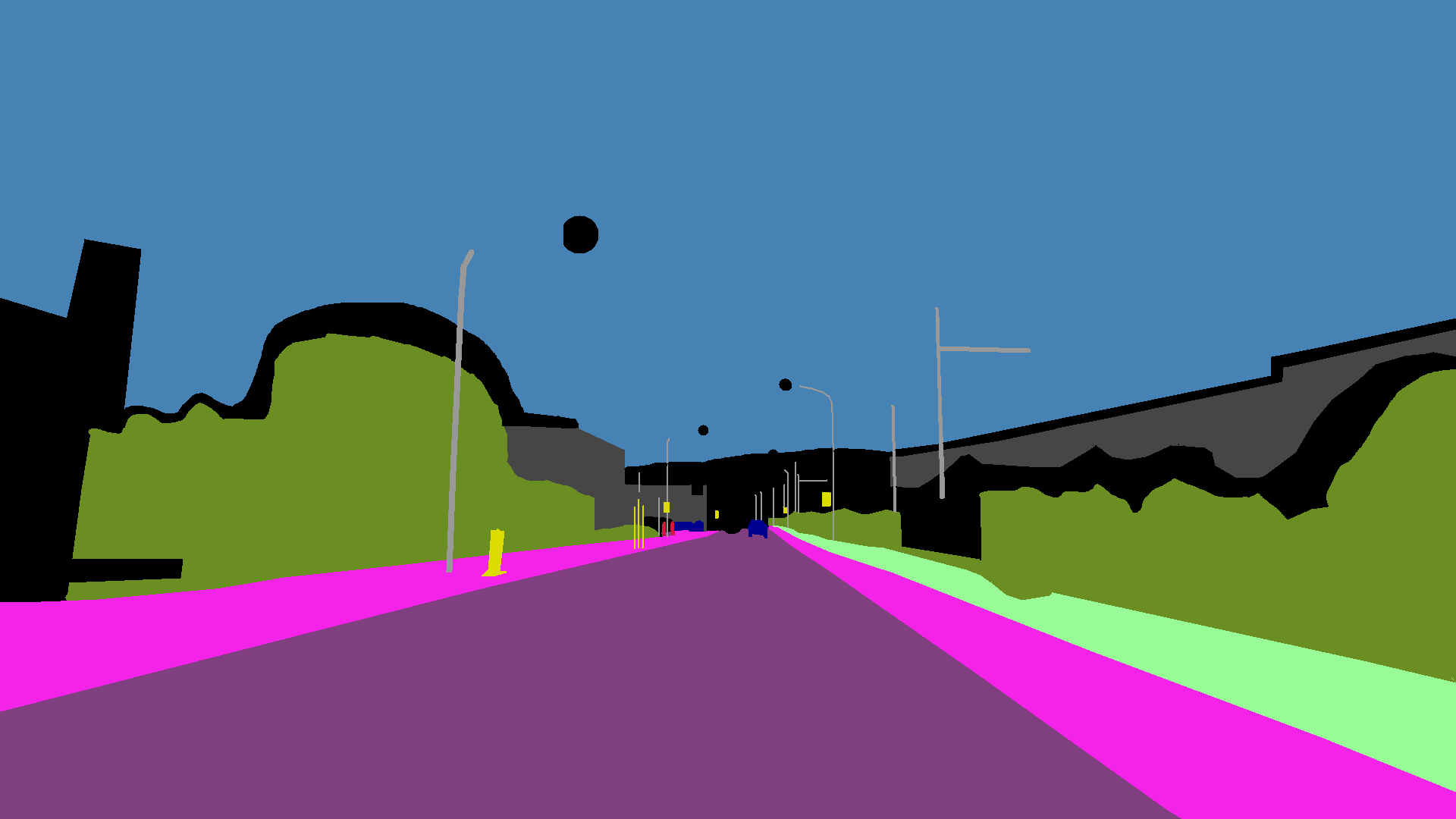}
        \end{minipage}
        \begin{minipage}{0.16\linewidth}
            \includegraphics[width=2.95cm]{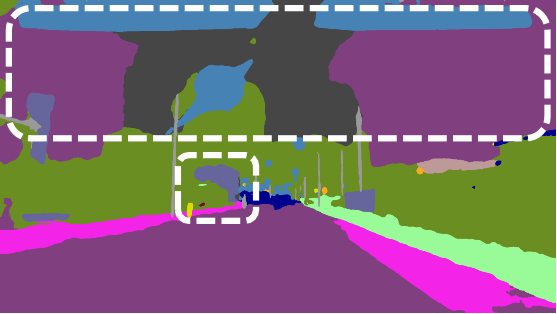}
        \end{minipage}
        \begin{minipage}{0.16\linewidth}
            \includegraphics[width=2.95cm]{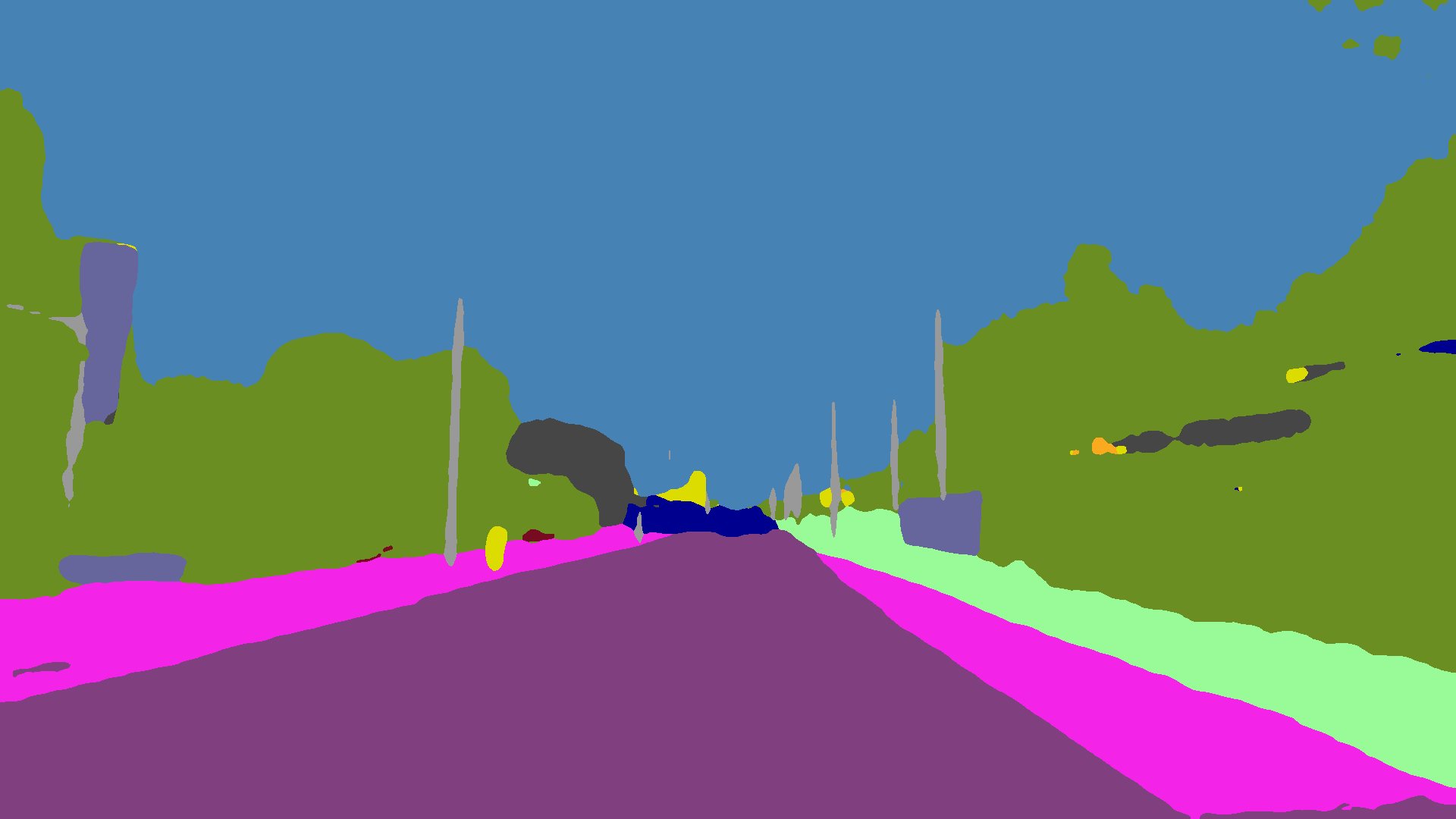}
        \end{minipage}
        \begin{minipage}{0.16\linewidth}
            \includegraphics[width=2.95cm]{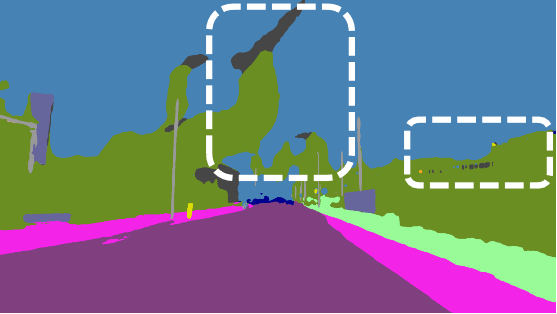}
        \end{minipage}
        \begin{minipage}{0.16\linewidth}
            \includegraphics[width=2.95cm]{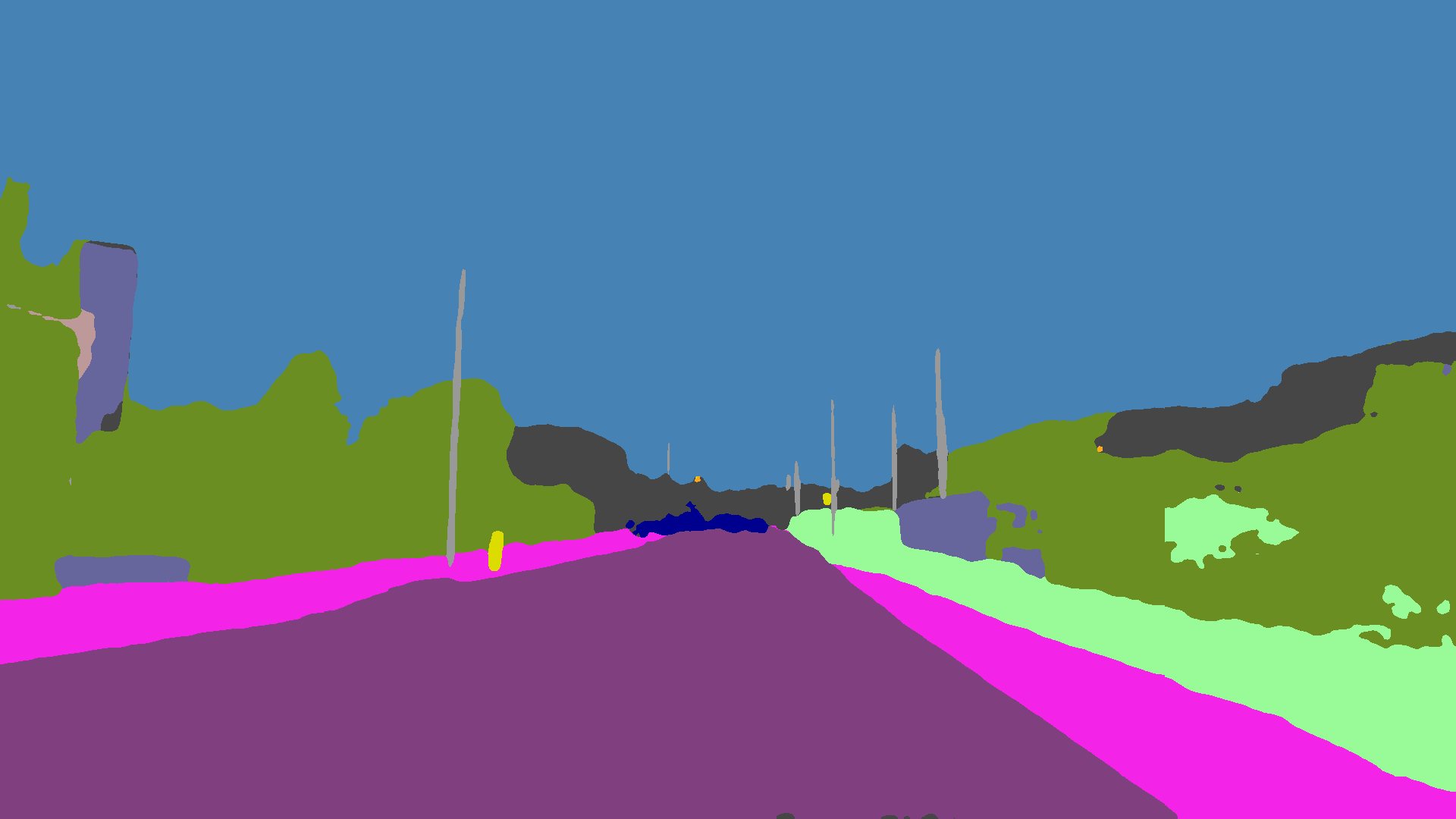}
        \end{minipage}
        \vskip 2pt
        \begin{minipage}{0.16\linewidth}
            \includegraphics[width=2.95cm]{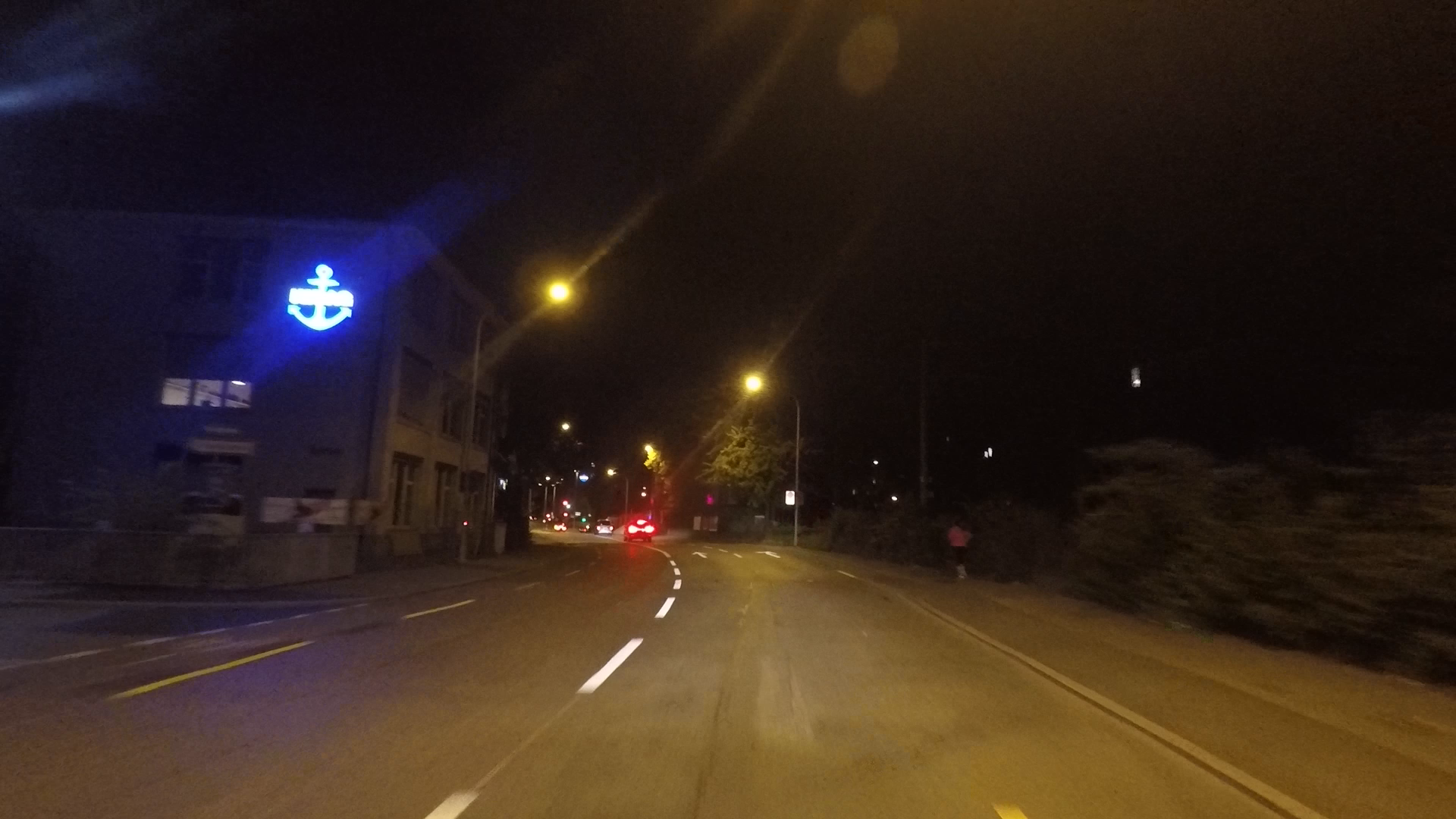}
        \end{minipage}
        \begin{minipage}{0.16\linewidth}
            \includegraphics[width=2.95cm]{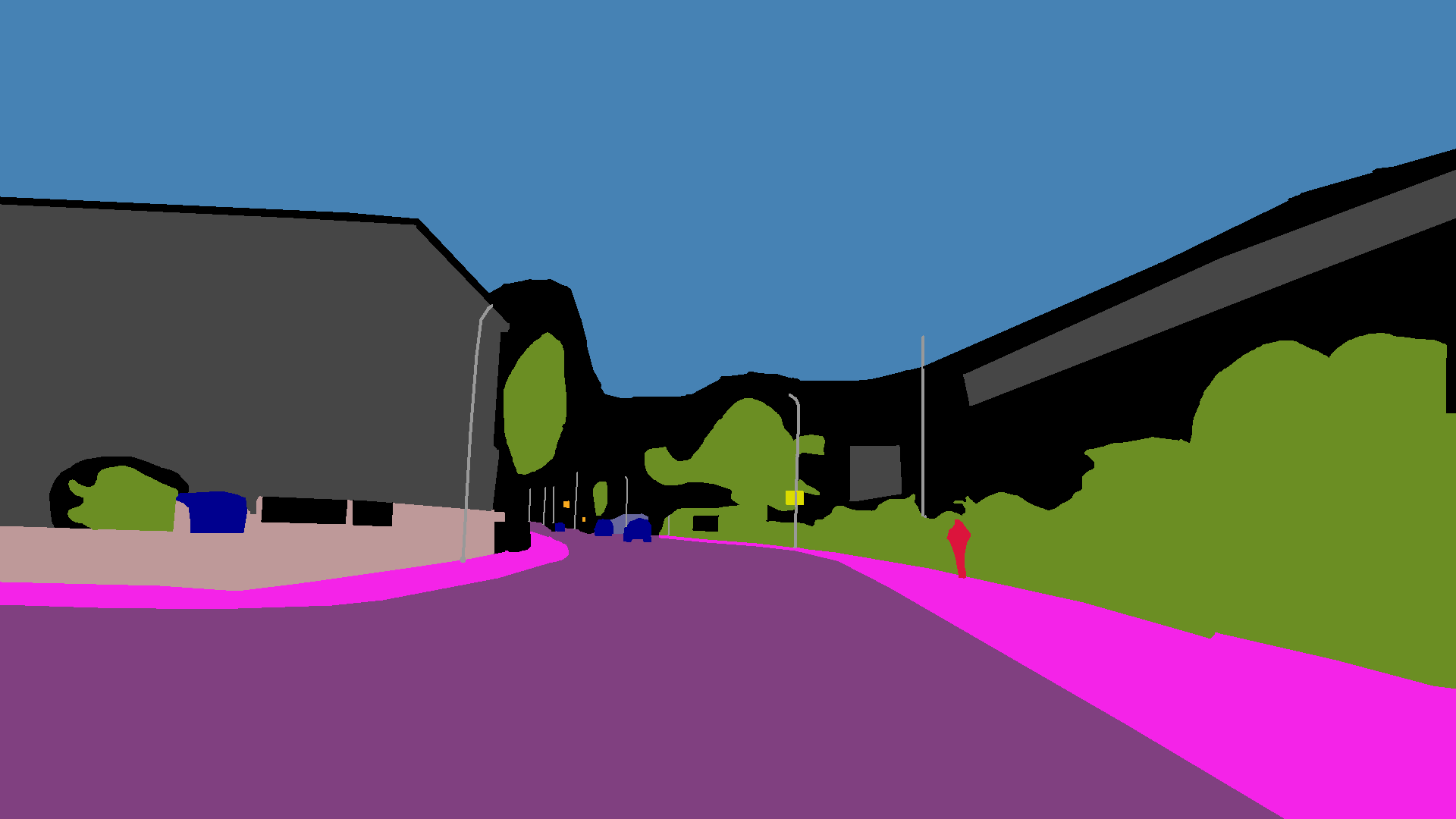}
        \end{minipage}
        \begin{minipage}{0.16\linewidth}
            \includegraphics[width=2.95cm]{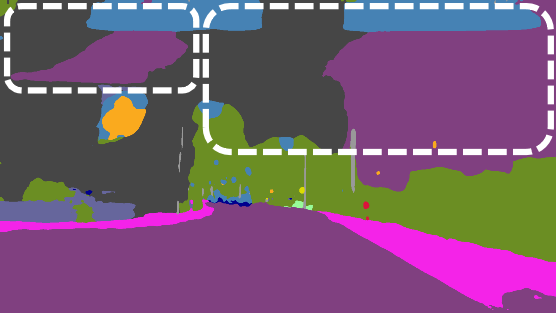}
        \end{minipage}
        \begin{minipage}{0.16\linewidth}
            \includegraphics[width=2.95cm]{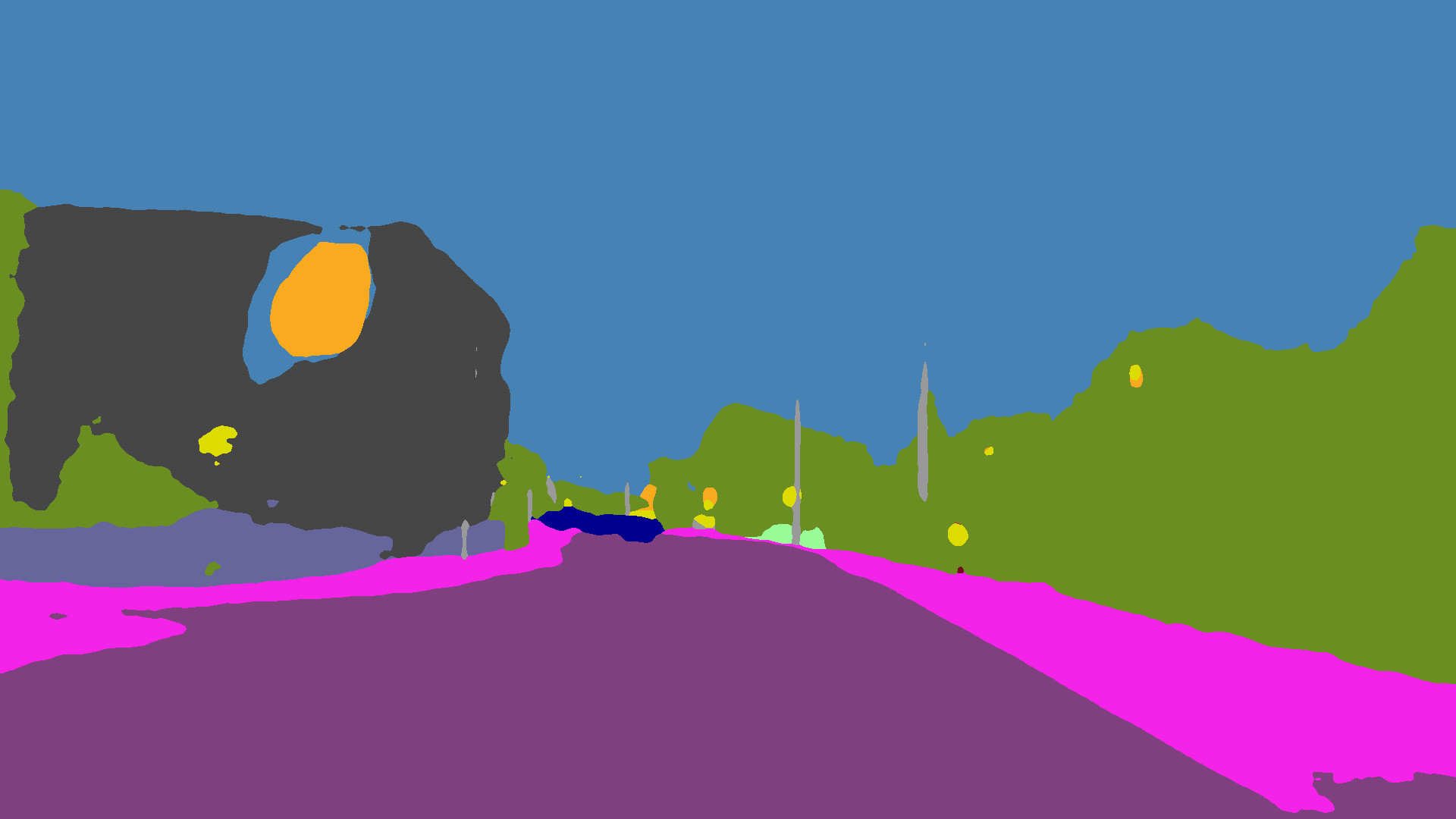}
        \end{minipage}
        \begin{minipage}{0.16\linewidth}
            \includegraphics[width=2.95cm]{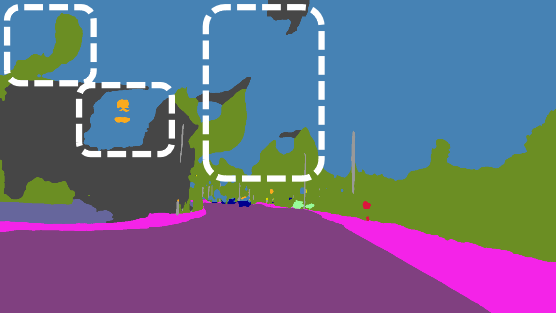}
        \end{minipage}
        \begin{minipage}{0.16\linewidth}
            \includegraphics[width=2.95cm]{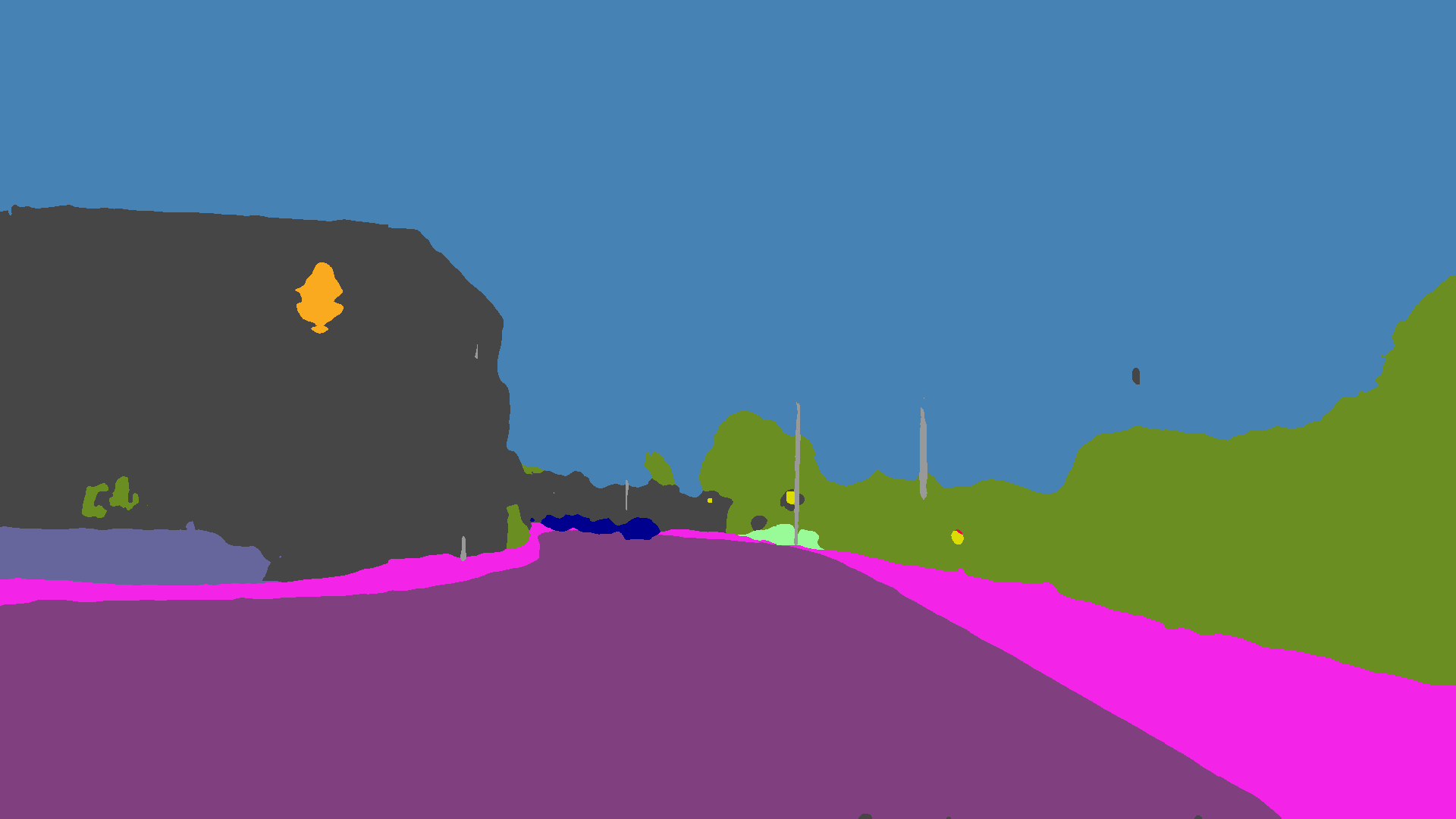}
        \end{minipage}
        \vskip 2pt
        \begin{minipage}{0.16\linewidth}
            \includegraphics[width=2.95cm]{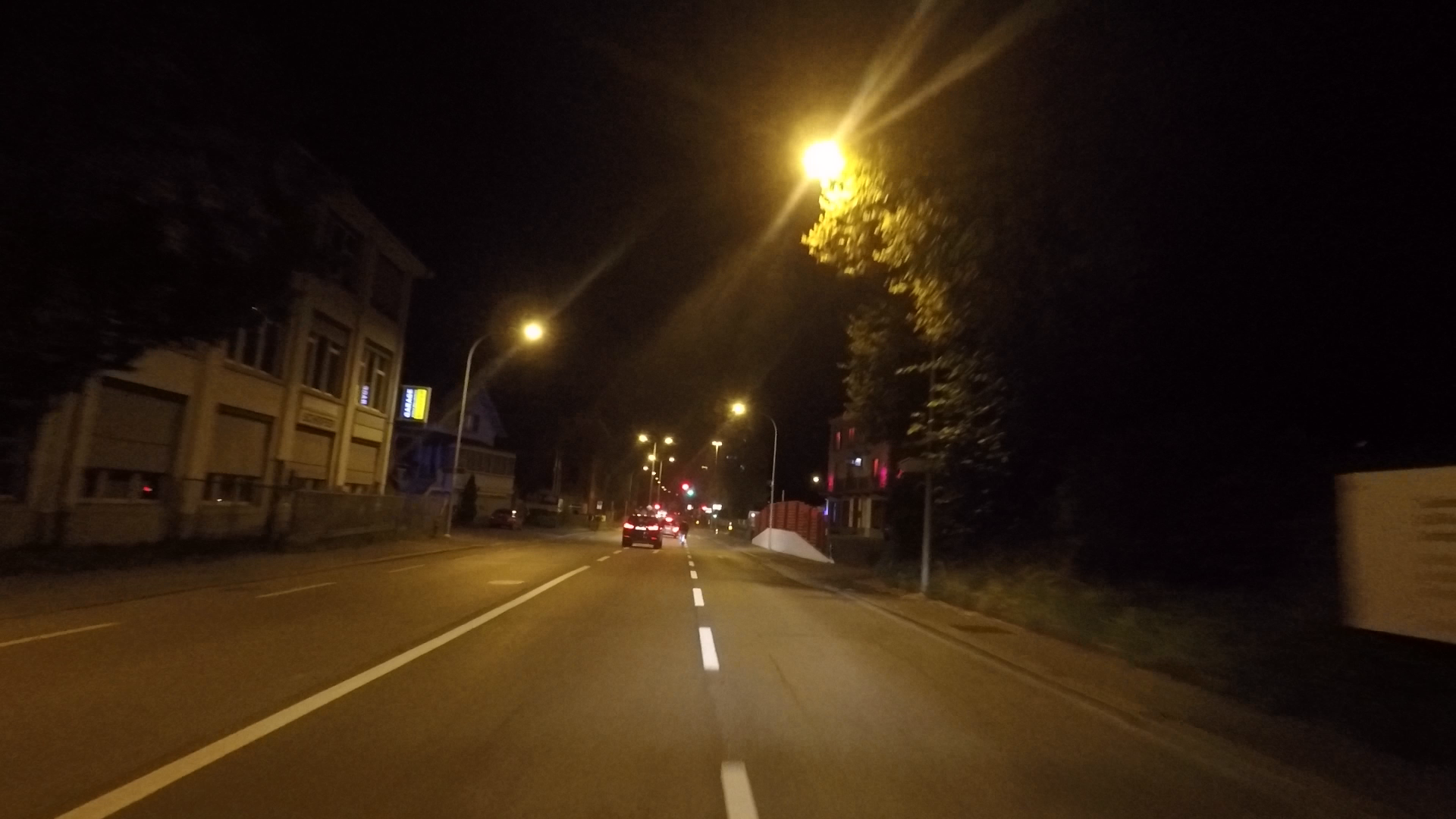}
        \end{minipage}
        \begin{minipage}{0.16\linewidth}
            \includegraphics[width=2.95cm]{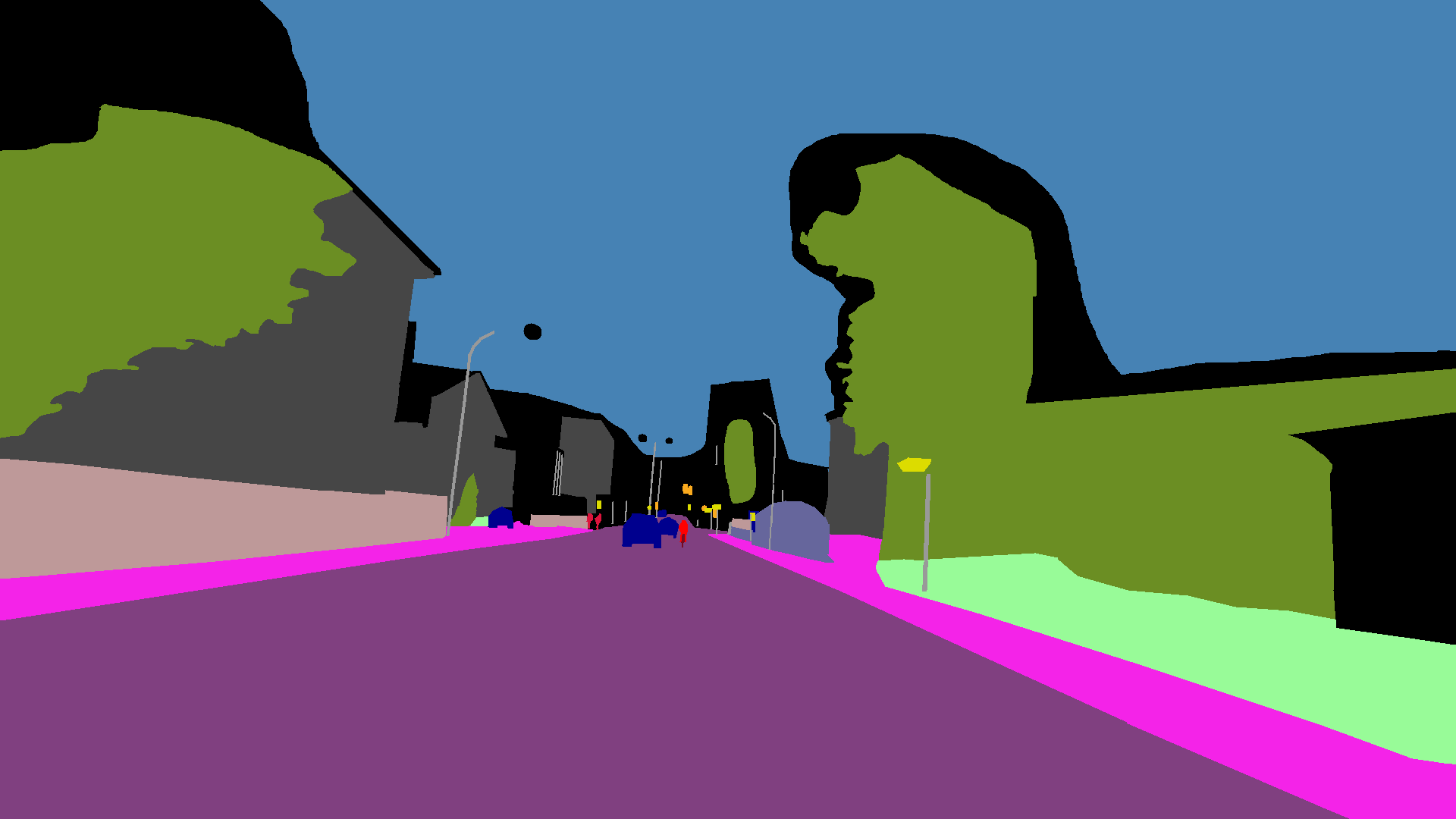}
        \end{minipage}
        \begin{minipage}{0.16\linewidth}
            \includegraphics[width=2.95cm]{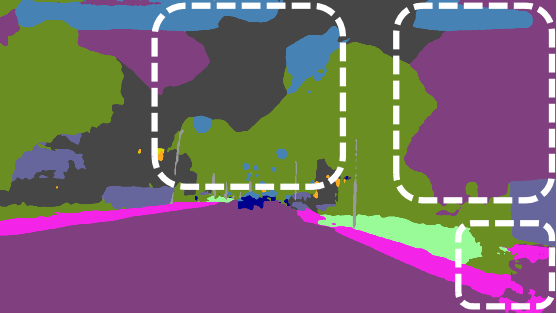}
        \end{minipage}
        \begin{minipage}{0.16\linewidth}
            \includegraphics[width=2.95cm]{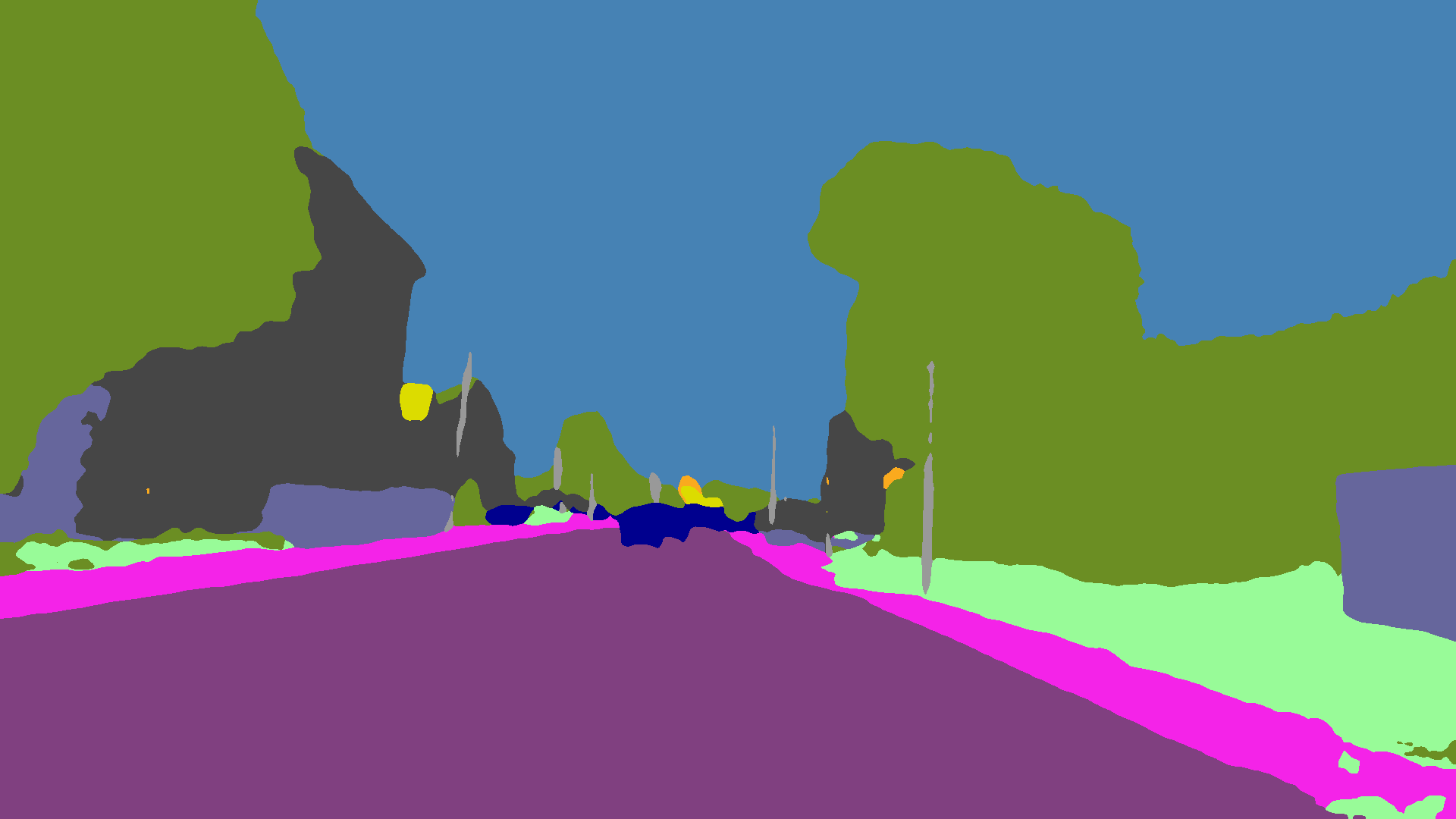}
        \end{minipage}
        \begin{minipage}{0.16\linewidth}
            \includegraphics[width=2.95cm]{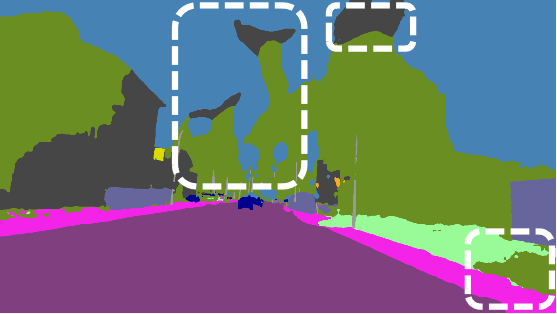}
        \end{minipage}
        \begin{minipage}{0.16\linewidth}
            \includegraphics[width=2.95cm]{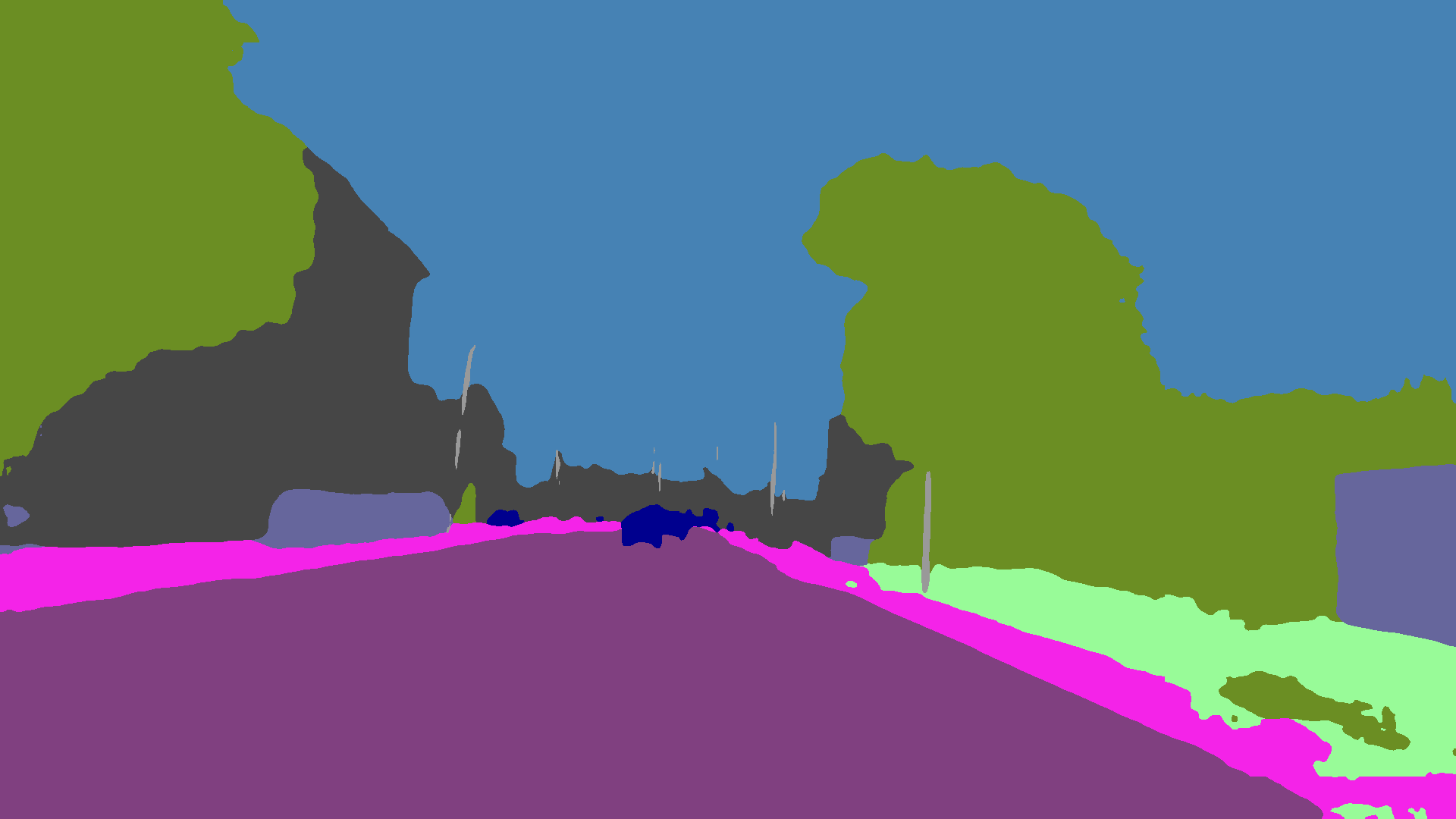}
        \end{minipage}
    \caption{Qualitative comparison of PIG with DAFormer and HRDA on Cityscapes to Dark Zurich. 
    In classes with a large domain gap between day and night such as sky, building, and vegetation, PIG achieved notable improvements.
    }
    \vspace{-5mm}
    \label{fig:cpmpare}
\end{figure*}

\textit{Qualitative Comparison:}
We compare the parsing results of night images using DAFormer, HRDA, PIG (DAFormer), and PIG (HRDA), as depicted in Fig. \ref{fig:cpmpare}. 
Regardless of whether based on DAFormer or HRDA, PIG exhibits a significant improvement in mitigating the impact of over/under exposure on model predictions. 
This improvement leads to a more reasonable distribution of predicted scene classes.
Furthermore, PIG demonstrates the ability to generate more accurate details, particularly in low-light junctions.
In UDA, distinguishing between the sky, building, and vegetation classes is challenging due to their similar features at night, despite being distinct during the day.
However, PIG addresses this issue by utilizing UDA and NFNet to handle classes with different domain similarities, thus resolving the problem of class confusion.

\subsubsection{Influences of Prompt Images}
Considering the direct impact of prompt images on the parsing of NFNet, we conduct a detailed analysis focusing on two main aspects: the number of prompt images and the number of classes contained in them.

\begin{figure}[t]
    \centering
    \includegraphics[width=0.48\textwidth]{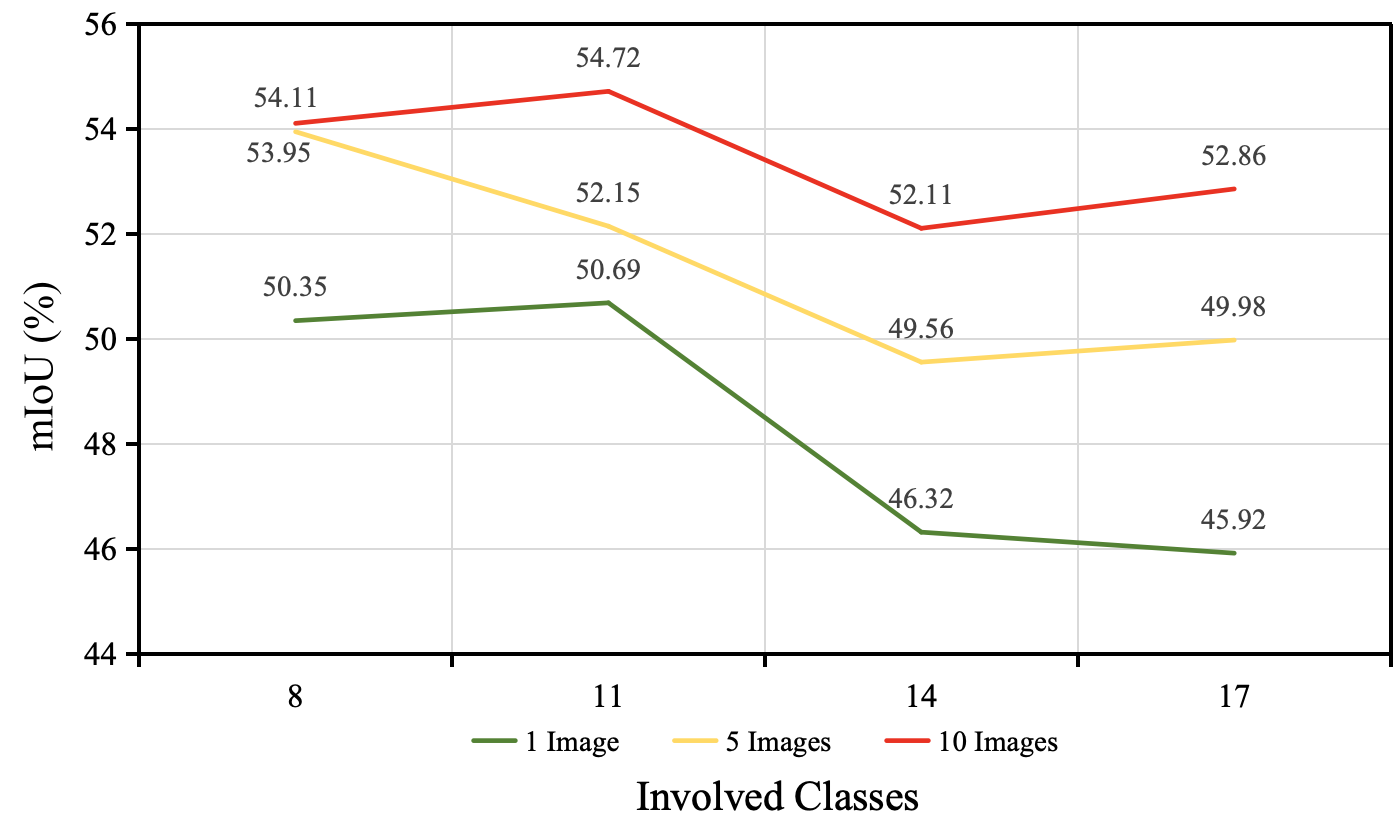}
    \vspace{-3mm}
    \caption{Parsing performance under different numbers of prompt images and the number of classes contained within them.
    }
    \vspace{-2mm}
    \label{fig:involve_class}
\end{figure}

\textit{The Number of Prompt Images:}
In order to investigate the impact of the number of prompt images, we conduct a parameter adjustment experiment using PIG (DAFormer), adapting from Cityscapes to Dark Zurich, as illustrated in Fig. \ref{fig:involve_class}. 
We employ 1, 5, and 10 night-time images as prompt images, respectively. 
The parsing results of PIG exhibit a gradual increase as the number of prompt images increased. 
Specifically, when utilizing 5 prompt images, and 8 classes contained, PIG achieves results 53.95 mIoU comparable to DAFormer 53.8 mIoU in Tab. \ref{tab:result}.
Moreover, with 10 prompt images, and 11 classes contained, PIG surpasses DAFormer by 0.9 mIoU.
Analyzing the average upward trend from 1 image to 5 images, and from 5 images to 10 images, we observe a decrease in performance improvement from +3.09 mIoU to +2.04 mIoU. 
This suggests that as the number of prompt images increases, the enhancement provided by PIG diminishes gradually in UDA.
We attribute this phenomenon to the design of the NFNet network. 
A limited set of labeled images enables NFNet to grasp the salient features of night-time scenes effectively. 
Even with an increase in data volume, NFNet primarily enhances the precision of detailed predictions, with limited impact on the prediction of salient features.
Consequently, it implies that PIG only requires a small number of prompt images to yield significant results.

\begin{figure}[!t]
    \centering
    \includegraphics[width=0.45\textwidth]{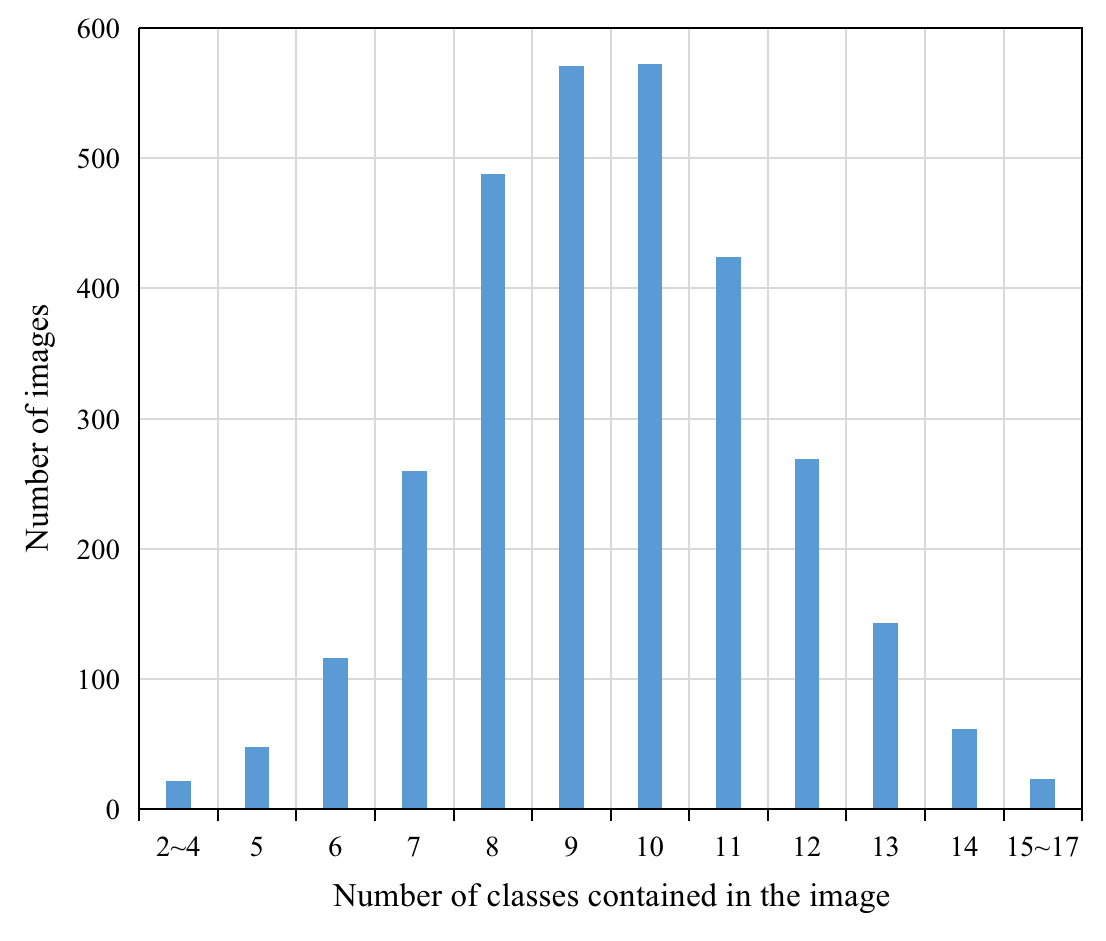}
    \vspace{-4mm}
    \caption{Statistics of the number of images containing different numbers of classes. The data come from 2998 training set images from NightCity.
    }
    \vspace{-5mm}
    \label{fig:nightcity_class}
\end{figure}

\textit{Involved Classes of Prompt Images:}
Given that each set contains no more than 10 prompt images, we are concerned about selecting the most suitable image as a prompt. 
We train our model, PIG (DAFormer), using a diverse range of images, adapting from Cityscapes to Dark Zurich.
Our observations indicate that a typical night-time scene image contains around 6 to 13 different classes, as depicted in Fig. \ref{fig:nightcity_class}.
To conduct our experiment, we collect four groups of prompt images, each containing a different number of classes: 8, 11, 14, and 17, as illustrated in Fig. \ref{fig:involve_class}.
The experimental results reveal that when the number of classes is 8 or 11, the parsing results show an average improvement of 3.2 mIoU compared to the results obtained with 14 or 17 classes.
Notably, when using only one prompt image, the gap increases to 4.4 mIoU, compared to using 5 or 10 prompt images.
These findings suggest that images containing too many classes are not suitable as prompt images for PIG.
Firstly, when an image contains a rich variety of classes, the distribution of objects becomes complex.
Complex scene distributions are often more difficult for the network to handle compared to simpler scenes. 
Secondly, we do not expect NFNet to achieve detailed predictions comparable to UDA using the limited night supervised information. 
Instead, NFNet's primary objective is to accurately predict the distribution of scenes that exhibit night-time features. 
A night image with an excessive number of classes will introduce redundant information that deviates from our intended focus.
In PIG, the UDA provides a large amount of supervised information that aids in predicting small objects, while the NFNet excels at learning difficult night features that are not easily captured in UDA.
Therefore, a simple scene with a few distributed small objects is more suitable as a prompt image for PIG.

\begin{figure*}[!t]
    \centering
    \includegraphics[width=1\textwidth]{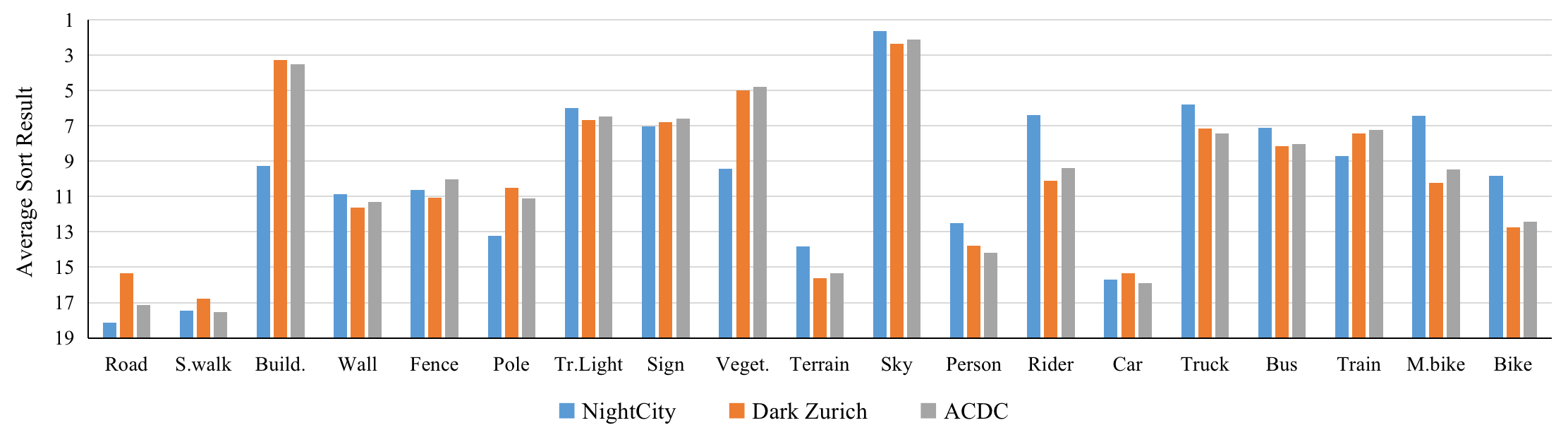}
    \vspace{-8mm}
    \caption{Classes sorting visualization of NightCity, Dark Zurich, ACDC in FDSG. Each night-time dataset is combined with the day-time dataset Cityscapes to form an average sort result of 5,000 day-night image pairs.
    }
    \vspace{-5mm}
    \label{fig:vis_lpips}
\end{figure*}

\subsubsection{Influence of the Class Weight in FDSG}
\begin{figure}[!t]
    \centering
    \includegraphics[width=0.48\textwidth]{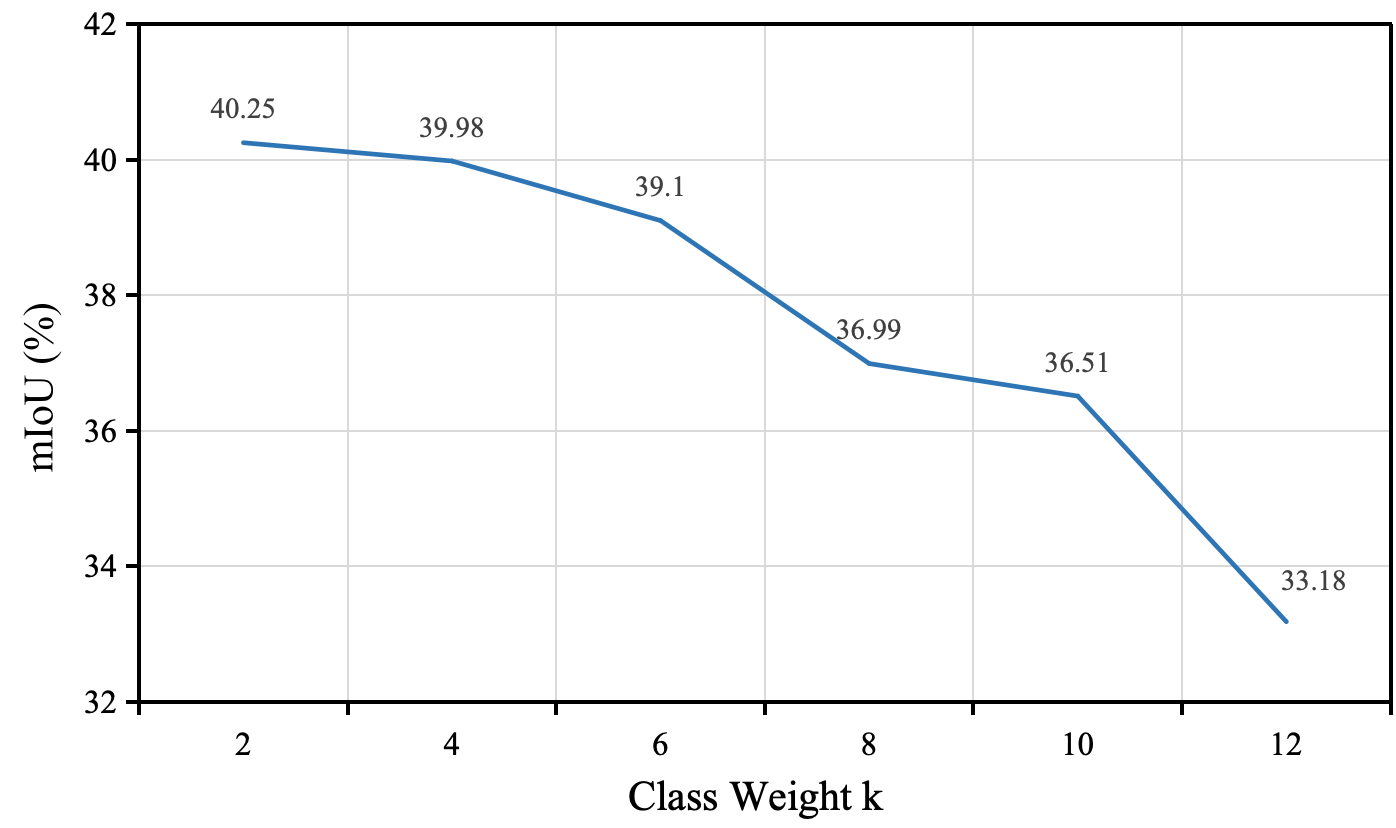}
    \vspace{-3mm}
    \caption{Parsing performance under different class weights $k$ in FDSG.
    }
    \vspace{-3mm}
    \label{fig:class_weight}
\end{figure}

The parameter $k$ in FDSG represents the weight assigned to the self-supervised pseudo-labels from both the UDA and NFNet.
A larger value of $k$ indicates a higher likelihood of the pseudo-label $\hat{y}^{T}_{fuse}$ being derived from the NFNet's prediction.
During our experiments, we adjust $k$ in PIG (DAFormer) from Cityscapes to NightCity to observe the influence of UDA and NFNet on the parsing results under various weights, as depicted in Fig. \ref{fig:class_weight}.
We observe that as $k$ increases from the first 2 classes to the first 6 classes, the parsing results decrease by an average of 0.58 mIoU with each increment. 
We attribute this decrease to normal fluctuations within the training process, which involves random image mixing. 
However, it is important to note that PIG consistently outperforms DAFormer (33.1 mIoU) in terms of overall results.
When $k$ increases from the first 6 classes to the first 12 classes, the parsing results exhibit a significant decline, with an average drop of 1.97 mIoU for each increment. 
This finding suggests that when PIG fuses pseudo-labels, submitting less than one-third of the classes for prediction by NFNet can lead to substantial improvements in parsing results. 
Furthermore, NFNet, which is trained exclusively on night images, proves superior to UDA when dealing with approximately six classes that possess distinct night features. 
In UDA, the supervision of day-time images can confuse the network when encountering classes with fewer domain similarities. 
By leveraging NFNet's predictions for these classes, PIG effectively avoids the adverse supervision from source domain images.

\subsubsection{Visualization of Classes Sorting in FDSG}
In order to investigate whether the FDSG module selects different classes when adapting the day-time domain to various night-time domains, we conduct visualizations of the NightCity, DarkZurich, and ACDC datasets after applying FDSG, as depicted in Fig. \ref{fig:vis_lpips}. 
We randomly select 100 day-time images and 1 night-time image and calculate the LPIPS results. 
We then average the ranking of each class to form a set of experiments.
Additionally, we randomly sample 50 night images for 50 groups of experiments and compute the average results. 
In the end, we combine each night-time dataset with the day-time dataset, creating 5,000 day-night image pairs for the experiment.
Fig. \ref{fig:vis_lpips} demonstrates that although FDSG may exhibit different class orderings for different night-time datasets, it generally maintains a consistent bias towards classes in the night-time domain. 
Specifically, FDSG tends to perceive fewer domain similarities in the areas of sky, traffic lights, and traffic signs, while perceiving more domain similarities in the areas of roads, sidewalks, and cars.

\subsubsection{Exploration of UDA and NFNet}
In our proposed PIG approach, the sole supervisory input for UDA is the labeled day-time image from the source domain, while for NFNet, it is derived from the labeled prompt image. 
To assess the performance of each network independently, we train both using this supervisory information and then conduct evaluations on a night-time test set, as illustrated in Tab. \ref{tab:uda_nf}.
As evidenced by the results in Tab. \ref{tab:uda_nf}, UDA networks, benefiting from extensive supervisory information, outperform NFNet significantly on the night-time test set. 
However, after implementing our PIG training method, we observe a substantial improvement in the parsing accuracy of the UDA network. 
This enhancement indicates that UDA and NFNet acquire distinct types of knowledge, which when combined, can have a synergistic effect. 
This finding also demonstrates that our method effectively leverages prior knowledge from day-time conditions and significant night-time features to enhance the UDA's parsing capabilities.

Additionally, to clearly demonstrate the distinctions between UDA and NFNet stemming from disparities in supervisory information, we compare the performance of NFNet, source-free UDA and general UDA methods, as illustrated in Tab. \ref{tab:source_free}.
Tab. \ref{tab:source_free} reveals that with only 10 prompt images as supervisory input, the parsing accuracy of the NFNet falls short of the source-free UDA method. 
It is important to note that the source-free UDA approach relies on an initial model that is pre-trained using source domain data, thereby gaining substantial supervisory information. 
Despite potential knowledge loss during subsequent training phases, the source-free UDA method generally outperforms NFNet. 
A comparison between source-free and general UDA methods shows that general UDA significantly outperforms source-free UDA on the test set, underscoring the importance of source domain supervision in domain adaptation processes.

\begin{table}[t]
\centering
\setlength{\tabcolsep}{2.5mm}
\caption{Comparison of UDA and NFNet in PIG (DAFomer).}
    \renewcommand{\arraystretch}{1.3}
    \begin{tabular}{@{}ccccc@{}}
        \hline
        Network & Training Dataset & Test Dataset & mIoU \\
    \hline
    UDA & Cityscapes & NightCity &  33.10\\
    NFNet & 10 Prompt Images & NightCity &  21.47\\
    PIG & Cityscapes \& Prompt Images & NightCity & 40.98\\
    \hline
    \end{tabular}
    \vspace{-3mm}
    \label{tab:uda_nf}
\end{table}

\begin{table}[t]
\centering
\setlength{\tabcolsep}{1.8mm}
\caption{Comparison of NFNet, Source-Free UDA (SF) and general UDA. $\dagger$ means that the dataset is used without human annotation.}
    \renewcommand{\arraystretch}{1.3}
    \begin{tabular}{@{}ccccc@{}}
    \hline
    Network & SF & Training & Test & mIoU \\
    \hline
    TTBN \cite{nado2020evaluating}& $\surd$ & Dark Zurich$\dagger$ & Dark Zurich & 28.00\\
    TENT \cite{wang2020tent} & $\surd$ & Dark Zurich$\dagger$ & Dark Zurich & 26.60\\
    AUGCO \cite{prabhu2022augmentation} & $\surd$ & Dark Zurich$\dagger$ & Dark Zurich & 32.40\\
    C-SFDA \cite{karim2023c} & $\surd$ & Dark Zurich$\dagger$ & Dark Zurich & 33.20\\
    \hline
    NFNet & $\surd$ & Prompt Images & Dark Zurich & 18.22 \\
    UDA (DAFormer) & & City. \& Dark Zurich$\dagger$ & Dark Zurich &  53.80\\
    PIG (DAFormer) & & City. \& Dark Zurich$\dagger$ & Dark Zurich & 54.72\\
    \hline
    \end{tabular}
    \vspace{-3mm}
    \label{tab:source_free}
\end{table}

\subsubsection{Exploration of Similarity Evaluation Methods}
In addition to the widely used LPIPS method, we also explore traditional methods for evaluating image similarity, such as PSNR and SSIM. 
These methods rely on the physical properties of the image, including luminance, color, and structure, to calculate the evaluation results. 
To conduct experiments on the NightCity dataset using PIG (DAFormer), we replace LPIPS with PSNR and SSIM, and the results are presented in Tab. \ref{tab:evaluation}.
The results clearly demonstrate that the performance assessed using SSIM and PSNR as evaluation methods falls notably short in comparison to LPIPS. 
We attribute this disparity to the fundamental distinctions between LPIPS and traditional methods. 
Traditional methods like SSIM and PSNR rely on objective physical properties for their calculations, thereby making them susceptible to variations in image content. 
This issue becomes particularly pronounced in scenes such as night-time cityscapes, characterized by low illumination and intricate structures, which introduce significant uncertainties into the evaluation process using traditional methods.
In contrast, LPIPS leverages a deep neural network pretrained on a vast dataset of images, enabling it to conduct image similarity assessments that effectively mitigate errors stemming from differences in image attributes.

\subsubsection{Experiment of Tuning Hyperparameters}
We set different weights for the hyperparameters in Eqn. (\ref{eqn_15}) and carry out PIG (DAFormer) experiments on the NightCity dataset, and the results are shown in Tab. \ref{tab:turning}.
We establish four groups of experiments by configuring the hyperparameters ($\lambda_1$, $\lambda_2$, $\lambda_3$) as follows: (0.5, 1, 1), (1, 0.5, 1), (1, 1, 0.5), and (1, 1, 1).
The results indicate a noticeable decrease in parsing performance as the hyperparameter $\lambda_1$ controlling UDA loss $\mathcal{L}_{UDA}$ decreases. 
We attribute this phenomenon to the diminished weight of effective supervision information in domain adaptation, underscoring the crucial role of supervision information from the source domain in aiding the model's understanding of scene features. 
Furthermore, within NFNet, the reduction in weight $\lambda_3$ of data augmentation loss $\mathcal{L}_{A}$ has a more pronounced impact on parsing outcomes compared to the reduction $\lambda_2$ in prompt supervision loss $\mathcal{L}_{P}$. 
This suggests that self-supervised training on diverse data can mitigate overfitting issues stemming from a limited number of prompt images.

\begin{figure}[!t]
    \centering
    \includegraphics[width=0.48\textwidth]{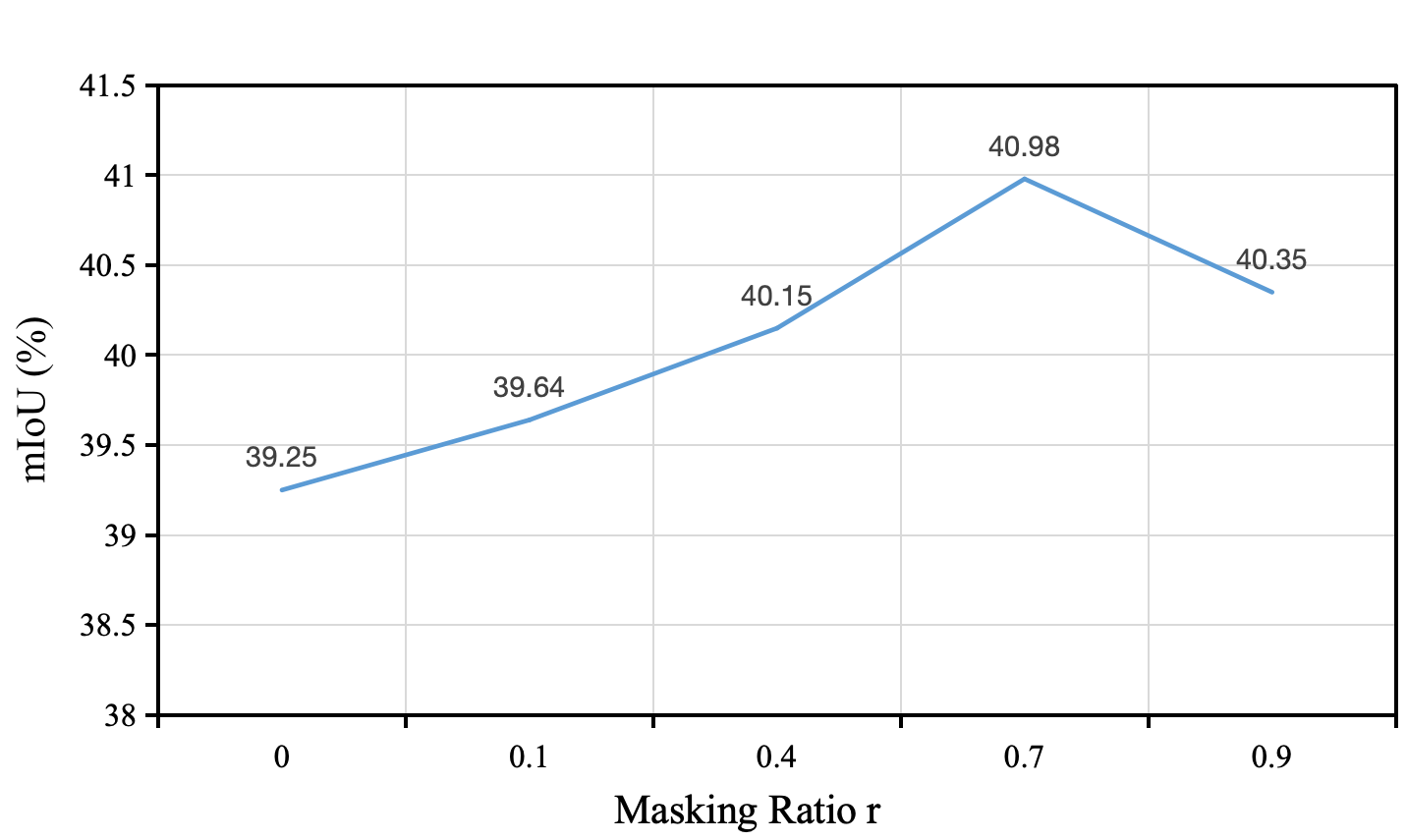}
    \vspace{-2mm}
    \caption{The parsing results of PIG under various masking ratios $r$.
    }
    \vspace{-3mm}
    \label{fig:mask_r}
\end{figure}

\begin{table}[!t]
\centering
\setlength{\tabcolsep}{6mm}
\caption{PIG performance under different domain similarity evaluation methods.}
    \renewcommand{\arraystretch}{1.3}
    \begin{tabular}{@{}ccc@{}}
        \hline
        Model & Evaluation Method & mIoU \\
    \hline
    DAFormer & - & 33.10 \\
    PIG (DAFormer) & SSIM & 33.90 \\
    PIG (DAFormer) & PSNR & 38.23 \\
    PIG (DAFormer) & LPIPS & 40.98 \\
    \hline
    \end{tabular}
    \vspace{-3mm}
    \label{tab:evaluation}
\end{table}

\begin{table}[t]
\centering
\setlength{\tabcolsep}{4.5mm}
\caption{Experiment of tuning hyperparameters in Eqn. (\ref{eqn_15}).}
    \renewcommand{\arraystretch}{1.3}
    \begin{tabular}{@{}ccccc@{}}
        \hline
        & $\lambda_1$ & $\lambda_2$ & $\lambda_3$ & mIoU \\
    \hline
    PIG (DAFormer) & 0.5 & 1 & 1 & 39.08 \\
    PIG (DAFormer) & 1 & 0.5 & 1 & 40.46 \\
    PIG (DAFormer) & 1 & 1 & 0.5 & 39.87 \\
    PIG (DAFormer) & 1 & 1 & 1 & 40.98 \\
    \hline
    \end{tabular}
    \vspace{-3mm}
    \label{tab:turning}
\end{table}

\subsubsection{Experiment of Tuning Masking Ratio $r$}
We conduct a parameter adjustment experiment on the NightCity dataset using PIG (DAFormer) to analyze the effects of masking ratio $r$, and the findings are presented in Fig. \ref{fig:mask_r}. 
We establish five experimental groups with $r$ values of 0, 0.1, 0.4, 0.7, and 0.9 respectively.
The results indicate a gradual increase in parsing performance as $r$ increases up to 0.7. 
However, when $r$ reaches a high value of 0.9, the parisng performance decreases. 
While masking aids in enhancing the model's ability to infer relationships between neighboring objects, an excessively high masking ratio hampers model training.

\subsection{Ablation Study}
To demonstrate the effectiveness of our proposed image training method and two data augmentation strategies, PMS and AMS, we conduct ablation studies on PIG (DAFormer) from Cityscapes to Dark Zurich, as presented in Tab. \ref{tab:ablation}.
Without any data augmentation strategy, the parsing results of PIG are significantly lower than those achieved by UDA methods.
Particularly, when supervised training is performed by directly adding prompt images to NFNet, the results decrease from 49.23 mIoU to 48.55 mIoU. 
This suggests that the inclusion of only a small amount of night supervision information does not directly improve the network's parsing accuracy of night scenes.
Instead, it can lead to overfitting on these night scenes, consequently reducing the overall model accuracy.
Regarding image mixing, we compare two approaches: cross-domain mixing (CDM) and prompt mixture strategy (PMS).
When only image mixing is used as the data augmentation method, the results for CDM and PMS are 51.99 mIoU and 52.01 mIoU, respectively, with no significant difference observed.
However, when AMS is introduced, the result for PMS increases to 54.72 mIoU, surpassing the 53.83 mIoU achieved by CDM. 
After training with masked images, the model can infer the class distribution of the unknown region from the information provided by adjacent objects. 
In comparison to CDM, PMS preserves the normal logical distribution found in night scenes and enhances the model's robustness at both the edge and center of the image. 
Consequently, when image mixing and mask training are combined, the mixing mode of PMS proves superior to that of CDM.

\begin{table}[!t]
    \centering
    \caption{Component ablation of the NFNet.}
    \renewcommand{\arraystretch}{1.3}
    \begin{tabular}{ccccccc}
    \hline
        & Target & Prompt & CDM & PMS & AMS & mIoU \\
        \hline
        1 & $\surd$ & - & - & - & - & 49.23\\
        2 & $\surd$ & $\surd$ & - &- & - & 48.55\\
        3 & $\surd$ & $\surd$ & $\surd$ & - & - & 51.99\\
        4 & $\surd$ & $\surd$ & - & $\surd$ & - & 52.01\\
        5 & $\surd$ & $\surd$ & $\surd$ & - & $\surd$ & 53.83\\
        6 & $\surd$ & $\surd$ & - & $\surd$ & $\surd$ & 54.72\\
    \hline
    \end{tabular}
    \vspace{-5mm}
    \label{tab:ablation}
\end{table}

\subsection{Discussion on UDA and SSDA}
In our research, we have identified a category of domain adaptation tasks, known as semi-supervised domain adaptation (SSDA), resembling our data settings.
However, it's crucial to note that our approach PIG differs from SSDA.
While SSDA aims to enable the network to learn features of the same class of objects in different domains simultaneously, our focus is on effectively separating feature differences between the two domains.
This distinction directly influences the reliance on labeled images from the target domain for both PIG and SSDA.
The SSDA model necessitates a considerable number of labeled images from the target domain (often exceeding 100) to attain a notable parsing advantage, whereas PIG remarkably enhances UDA with a minimal requirement of only 10 images.
By adopting PIG, we can develop a cost-effective competitive scene parsing model.

Additionally, we investigate the effect of domain adaptation using SSDA with very few labeled target domain images, and we compare it with UDA, as summarized in Tab. \ref{tab:UDA&SSDA}. 
The results in Tab. \ref{tab:UDA&SSDA} indicate that when the number of labeled images in the target domain is less than 50, the SSDA method does not show a significant advantage over UDA, but it does increase the labeling cost.
This finding suggests that SSDA requires a sufficient amount of labeled data from the target domain to demonstrate its effectiveness.
Therefore, we select UDA as the baseline for PIG for two primary reasons.
Firstly, to minimize labeling costs as much as possible.
Secondly, UDA achieves better parsing accuracy even without labeled images from the target domain, indicating that minimal use of supervision information from the target domain might interfere with the segmentation network's ability to learn object information. Our goal is to preserve similar features between the two domains while separating features with obvious differences, and UDA outperforms SSDA in retaining these similar features.

\begin{table}[!t]
    \centering
    \caption{Comparison between UDA and SSDA in different cases with very few target domain labeled images. G and C stand for GTA5 and Cityscapes.}
    \renewcommand{\arraystretch}{1.3}
    \begin{tabular}{c|lccc}
    \hline
        Type & Method & Target-labels & G $\rightarrow$ C (mIoU)\\
        \hline
        \multirow{8}{0.7cm}{SSDA} & \multirow{3}{1.9cm}{LabOR \cite{shin2021labor}} & 20 & 61.10\\
        ~ & ~ & 45 & 64.60\\
        ~ & ~ & 65 & 66.60\\
        \cline{2-4}
        ~ & \multirow{2}{1.9cm}{RIPU \cite{xie2022towards}} & 65 & 69.60\\
        ~ & ~ & 150 & 71.20\\
        \cline{2-4}
        ~ & \multirow{3}{1.9cm}{ILM-ASSL \cite{guan2023iterative}} & 30 & 70.00\\
        ~ & ~ & 65 & 75.00\\
        ~ & ~ & 150 & 76.10\\
        \hline
        \multirow{3}{0.6cm}{UDA} & DAFormer \cite{hoyer2022daformer} & - & 68.30\\
        ~ & SePiCo \cite{xie2023sepico} & - & 70.30\\
        ~ & HRDA \cite{hoyer2022hrda} & - & 73.80\\
    \hline
    \end{tabular}
    \vspace{-3mm}
    \label{tab:UDA&SSDA}
\end{table}

\section{Conclusion}
In this work, we propose Prompt Images Guidance (PIG) to improve unsupervised domain adaptation (UDA) without relying on day-night image pairs. 
Our approach effectively handles classes with fewer domain similarities between the source and target domains.
We introduce the Night-Focused Network (NFNet), which learns night-specific features from night images using the same architecture as UDA.
We design a Pseudo-label Fusion via Domain Similarity Guidance (FDSG), a fusion method that combines predictions from UDA and NFNet based on different domain similarities for each class. 
This fusion generates high-quality pseudo-labels to supervise the entire model.
To enhance the robustness of NFNet and prevent overfitting to prompt images, we propose two data augmentation strategies: the Prompt Mixture Strategy (PMS) and the Alternate Mask Strategy (AMS).
The PMS creates task prompt images by randomly combining vertically segmented target domain images and prompt images. 
NFNet learns to parse the remaining half of the night images using half of the prompt images. 
The AMS enables NFNet to leverage contextual information from adjacent objects in the image and preserve small object details.
We adapt PIG from Cityscapes to NightCity, NightCity+, Dark Zurich, and ACDC datasets, resulting in improvements in parsing accuracy for UDA.

\bibliographystyle{ieeetr}

\end{document}